\documentclass{article}
\usepackage{arxiv}
\usepackage{amsmath,amsfonts}
\usepackage{algorithmic}
\usepackage{algorithm}
\usepackage{array}
\usepackage[caption=false,font=normalsize,labelfont=sf,textfont=sf]{subfig}
\usepackage{textcomp}
\usepackage{stfloats}
\usepackage{url}
\usepackage{booktabs} 
\usepackage{colortbl} 
\usepackage{xcolor} 
\usepackage{multirow}
\usepackage{verbatim}
\usepackage{graphicx}

\usepackage{balance}
\graphicspath{{fig/}} 

\usepackage{nomencl}
\makenomenclature
\usepackage{multicol}

\usepackage[utf8]{inputenc} 
\usepackage[T1]{fontenc}    
\usepackage{hyperref}       
\usepackage{url}            
\usepackage{booktabs}       
\usepackage{amsfonts}       
\usepackage{nicefrac}       
\usepackage{microtype}      
\usepackage{lipsum}		
\usepackage{graphicx}
\usepackage{natbib}
\usepackage{doi}

\renewcommand{\shorttitle}{\textit{arXiv} Template}

\hypersetup{
pdftitle={A template for the arxiv style},
pdfsubject={q-bio.NC, q-bio.QM},
pdfauthor={David S.~Hippocampus, Elias D.~Striatum},
pdfkeywords={First keyword, Second keyword, More},
}

\newtheorem{proof}{Proof}[section]
\newtheorem{theorem}{\bf Theorem}

\begin{document}

\title{HOPS: High-order Polynomials with Self-supervised Dimension Reduction for Load Forecasting}

\author{Pengyang Song,
    Han Feng,
    Shreyashi Shukla,
    Jue Wang,
    and Tao Hong
\thanks{This work was supported by the National Social Science Foundation of China (No. 22VRC056) and the National Natural Science Foundation of China (No. 72271229, No. 71771208).} 
\thanks{Pengyang Song, Han Feng and Jue Wang are with (1) AMSS Center for Forecasting Science, Chinese Academy of Sciences, Beijing, 100190, China, and (2) MOE Social Science Laboratory of Digital Economic Forecasts and Policy Simulation, University of Chinese Academy of Sciences, Beijing, 100190, China. Correspond to wjue@amss.ac.cn.}
\thanks{Shreyashi Shukla and Tao Hong are with the Industrial and Systems Engineering Department, University of North Carolina at Charlotte, Charlotte, NC 28223 USA (e-mail: hongtao01@gmail.com).}
}

\maketitle

\begin{abstract}
Load forecasting is a fundamental task in smart grid. Many techniques have been applied to developing load forecasting models. Due to the challenges such as the Curse of Dimensionality, overfitting, and limited computing resources, multivariate higher-order polynomial models have received limited attention in load forecasting, despite their desirable mathematical foundations and optimization properties. 
In this paper, we propose low rank approximation and self-supervised dimension reduction to address the aforementioned issues. To further improve computational efficiency, we also utilize a fast Conjugate Gradient based algorithm for the proposed polynomial models. 
Based on the load datasets from the ISO New England, the proposed method high-order polynomials with self-supervised dimension reduction (HOPS) demonstrates higher forecasting accuracy over several competitive models. 
Additionally, experimental results indicate that our approach alleviates redundant variable construction, 
achieving better forecasts with fewer input variables.
\end{abstract}


\begin{multicols}{2}
\printnomenclature
\end{multicols}

\nomenclature{$n$}{The dimensionality of input variables.}%
\nomenclature{$d$}{The order of a polynomial.}%
\nomenclature{$\textbf{sum}$}{The point-by-point sum of tensors.}%
\nomenclature{$\boldsymbol{W_i}$}{The coefficient tensor of $i$-order monomials. When labeling subscripts as $(\boldsymbol{W_i})_{k_i \times k_i \cdots k_i}$ (or $(\boldsymbol{W_i})_{k_i}$), the subscript represents the shape of the tensor.}
\nomenclature{$m$}{The number of samples.}%
\nomenclature{$(x_{1\times n})_p$}{The feature vector corresponding to $p$-th sample.}
\nomenclature{$\boldsymbol{y}_p$}{The label corresponding to $p$-th sample.}
\nomenclature{$\|\cdot\|_F$}{The Frobenius norm.}
\nomenclature{$\ast$}{The Hadamard (element-wise) product.}%
\nomenclature{$\otimes$}{Tensor product (outer product). }%
\nomenclature{$L_{n \times k}$}{The linear dimension reduction matrix.}
\nomenclature{$\tilde{X}_{m \times k}$}{Obtained by $X_{m \times n} \cdot L_{n \times k}$.}
\nomenclature{$\tilde{x}_{1 \times k}$}{Obtained by $x_{1\times n} \cdot L_{n \times k}$.}
\nomenclature{$D_{n \times n}$}{The linear dimensionality reduction matrix.}
\nomenclature{$X_{m \times n}$}{A matrix obtained by arranging $x_{1 \times n}$ in rows.}
\nomenclature{$k_i$}{The embedding dimension of the feature variable corresponding to the i-order terms.}
\nomenclature{$\widetilde{X_i}$}{$\bigotimes^{i}(\widetilde{x}_{1 \times k_i})$ = $\bigotimes^{i}(x_{1 \times n} \cdot L_{n \times k_i})$.}
\nomenclature{$H_t$}{Class variable, 24 hours per day.}
\nomenclature{$W_t$}{Class variable, 7 days of a week.}
\nomenclature{$M_t$}{Class variable, 12 months of a year. }
\nomenclature{$Trend_t$}{A chronological trend.}
\nomenclature{${DEW}_t$}{The dew point temperature.}
\nomenclature{$T_t$}{The temperature (dry bulb temperature).}
\nomenclature{${RH}_t$}{relative humidity (\%).}
\nomenclature{$S_t$}{Dummy variable indicating Jun.-Sep. of the year.}
\nomenclature{$RHS_t$}{Cross effect of relative humidity and summer.}
\nomenclature{$RHS_t^2$}{Cross effect of relative humidity square and summer.}
\nomenclature{$\widetilde{T}_{t,b}$}{The daily moving average temperature of the $b$-th day}

\section{Introduction}
Electric load forecasting plays a crucial role in the contemporary energy industry. The accuracy of load forecasting models significantly affects various aspects of smart grid operations and planning, such as demand response, customer profiling, and upgrade of equipment. Many review articles can be found in the load forecasting literature \cite{weron2006modeling,hong2016probabilistic,hippert2001neural,wang2018review}. Various techniques have been proposed for load forecasting, such as multiple linear regression (MLR) \cite{charlton2014refined,hong2013long}, 
artificial neural networks (ANN) \cite{hippert2001neural,dimoulkas2019neural,xie2016relative}, and
support vector regression (SVR) \cite{chen2004load,luo2023robust}. 
Three Global Energy Forecasting Competitions, 
including GEFCom2012 \cite{hong2014global}, GEFCom2014 \cite{XIE20161012} and 
GEFCom2017 \cite{hong2019global}, have stimulated the enthusiasm of the researchers across multiple disciplines, leading to further proposals of many effective methods. Several methods can been found from the reports of multiple winners of the three aforementioned load forecasting competitions, such as outlier detection and data cleaning prior to forecasting \cite{charlton2014refined, XIE20161012}, 
least absolute shrinkage and selection operator (LASSO)\cite{ziel2016lasso}, forecast combination \cite{xie2016gefcom2014, haben2016hybrid, nowotarski2016improving}, and 
several machine learning techniques such as XGBoost and Gradient Boosting Machines
\cite{hong2019global, lloyd2014gefcom2012, taieb2014gradient}. 

From the perspective of model fitting, the models used by the contestants, even to the extend of the load forecasting community, can be roughly divided into two groups: statistical models, such as Multiple Linear Regression (MLR), and machine learning models, such has XGBoost and Artificial Neural Networks (ANN). There is no definitive distinction in forecasting accuracy between these two groups. Each has its own advantages and disadvantages. Statistical models are typically characterized by good interpretability, which allows researchers and practitioners to explain, refine, and defend their models. Machine learning models, on the other hand, tend to excel in fitting, which enables the forecasters to capture more intricate functional relationships between load and its driving factors. 

Given these considerations, there is a clear motivation to identify a model that synthesizes the strengths of both approaches. Ideally, such a model should possess strong mathematical properties that facilitate improvements while offering superior fitting capabilities 
compared to traditional statistical models.
In this regard, multivariate higher-order polynomials emerge as a promising candidate. 
However, despite the extensive use of sophisticated models in load forecasting competitions and related literature, the multivariate higher-order polynomial model has received comparatively little attention. A notable example is a family of interaction regression models with many third-order polynomial terms, which lead to hundreds of parameters to estimate for each model \cite{wang2016electric}.

Arguably, polynomials are more frequently encountered in mathematics and statistics literature than in engineering-focused articles. We conjecture that there are three main reasons. The first one is The Curse of Dimensionality. The multivariate higher-order polynomial model inherently encounters the challenge of the Curse of Dimensionality, which results in excessively high computational complexity. For instance, when utilizing an $n$-dimensional $d$-order polynomial model for fitting, we require at least $ \binom{d+n}{d}$ optimizable parameters to fully define the model. Even with sufficient computer memory to store these parameters, the time required to solve the model can be prohibitively long. In \cite{luo2023robust}, the authors attempted to employ a quadric surface (an $n$-dimensional $2$-order polynomial model) to enhance the Support Vector Regression (SVR) model. However, due to the Curse of Dimensionality, they were forced to limit their analysis to only the diagonal elements of the quadratic term parameter matrix, which significantly diminished the model's fitting capability compared to a fully parameterized polynomial model.

The second reason is overfitting. Although an increased number of parameters allows for a broader hypothesis space and more alternative functions for fitting training sets, it can also lead to severe overfitting. As the order of the polynomial rises, the number of parameters grows exponentially, making overfitting an inevitable consequence. As shown in \cite{wang2016electric}, as more third-order polynomials of lag temperatures are added to the model, the forecast error decreases at first, and then increases when even more terms are added. 

The third reason is limited computing power from both software and hardware. Load forecasting models based on high-order polynomials may include hundreds of parameters, which may require a long time for model selection and various tests. In addition, a large regression model may present numerical issues, so that many software packages may not be able to provide accurate estimation results. In fact, the primary reason the models proposed in \cite{wang2016electric} did not include temperatures of higher than the third order was due to the limitations in software and hardware at the time. 

In light of these shortcomings, we aim to enhance the basic multivariate higher-order polynomial model to mitigate the impact of these issues while preserving the model's overall effectiveness. Simultaneously, we also notice that in the models frequently utilized for load forecasting tasks 
\cite{hahn2009electric,weron2006modeling, charlton2014refined,hong2013long,xie2016relative,xie2017variable,hong2010short,wang2016electric}, 
researchers usually improve the forecasting accuracy by constructing complex interaction variables. 
Very often significant experience or expertise is required to select the appropriate combination 
of variables via a large number of empirical comparisons as done in \cite{xie2017variable}. 
Therefore, we also hope to utilize the multivariate higher-order polynomial models while eliminating the process of constructing complex interaction variables.

Our contributions can be summarized in the following three aspects. First, drawing on concepts of self-supervised dimension reduction and low-rank approximation, we introduce a dimension reduction approach for the high-order terms of the polynomial model. This method enables us to achieve superior forecasting accuracy while significantly reducing the number of required parameters. Secondly, we introduce a Conjugate Gradient-based (CG) optimization algorithm, which facilitates a more efficient and rapid solution to our improved polynomial model. Thirdly, our approach eliminates the need for complex interaction variable construction. Through numerical experiments, we demonstrate that
our model can achieve superior performance compared to its counterparts even with fewer variables.

All numerical experiments are conducted on datasets from the ISO New England. The results demonstrate that the proposed method high-order polynomials
with self-supervised dimension reduction (HOPS) exhibits significant advantages over the benchmark and several competitive comparison models.

The remainder of this paper is organized as follows:
Section II presents the fundamental methodology, including the mathematical formulation of the basic multivariate higher-order polynomial model
and the self-supervised dimensionality reduction method. Related properties are also discussed in this section.
Section III details the algorithm designed to accelerate computation for HOPS.
Section IV outlines the experimental configuration, results, and analysis.
Section V concludes our work and discusses potential future research directions.
Finally, the Appendix includes proofs and statements related to the properties mentioned in Section II.

\section{Higher-order Polynomial Model and Self-supervised Dimension Reduction}

\subsection{The Definition of $n$-dimensional $d$-order Polynomial }

\begin{figure}[!t]
  \centering
  \includegraphics[width=4in]{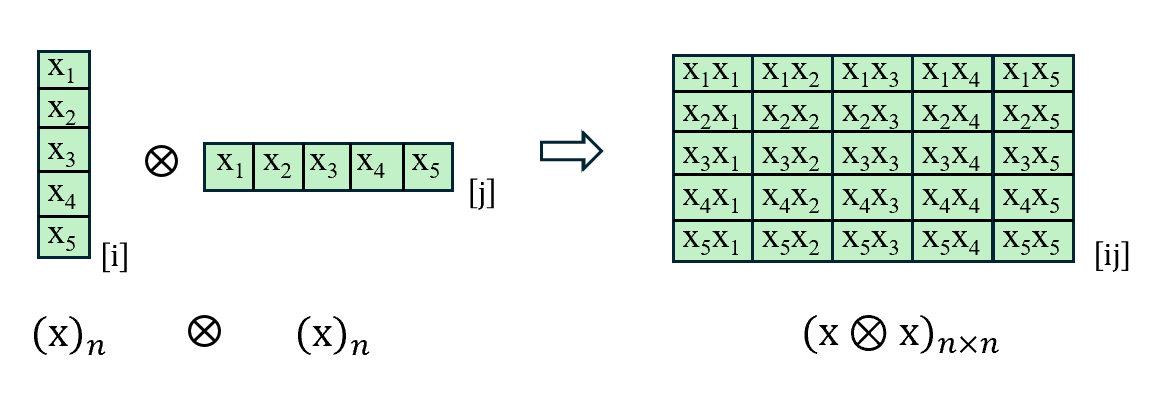}
  \includegraphics[width=4in]{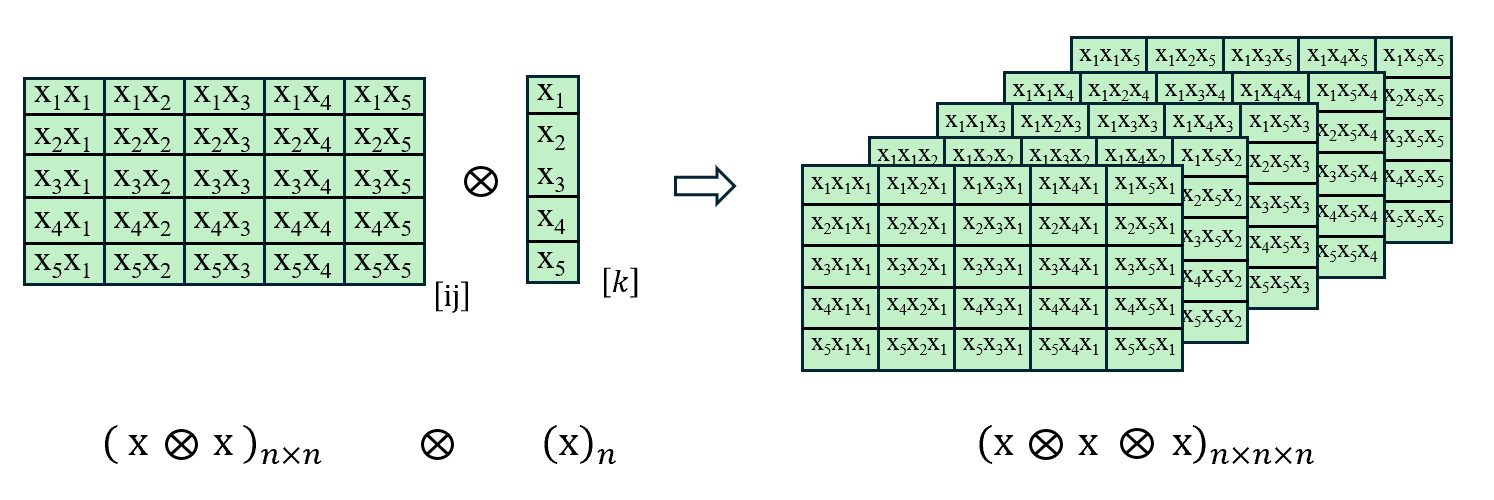}
  \caption{The tensor product of $x$ and $x$,  $x \otimes x$ and $x$.}
  \label{tensor1}
  \end{figure}

\begin{figure}[!t]
\centering
\includegraphics[width=6in]{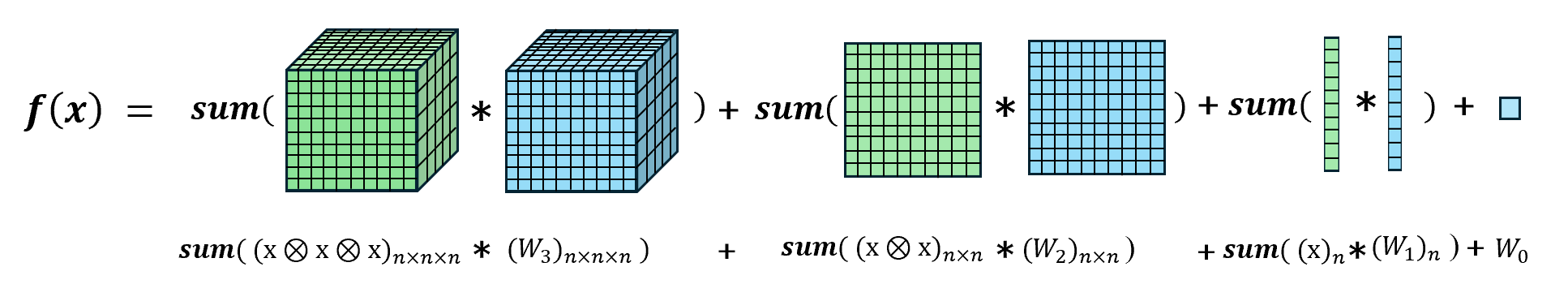}
\caption{An $n$-dimensional $3$-order polynomial.}
\label{model1}
\end{figure}

For an input feature vector $x_{1\times n}$ with $n$ input variables, the $n$-dimensional $d$-order polynomial $f$ is expressed as:
\begin{align} \label{multipolynomial}
&f\left(x_{1\times n}, \boldsymbol{W_0},\boldsymbol{W_1},\cdots \boldsymbol{W_d}\right) \\=
&\textbf{sum}(\boldsymbol{W_0}*X_0) \notag 
+ \textbf{sum}(\boldsymbol{W_1}*X_1) +\cdots, 
+ \textbf{sum}(\boldsymbol{W_d}*X_d) ,
\end{align}
where $\textbf{sum}()$ denotes the point-by-point sum of tensors. $\boldsymbol{W_i} \in \mathbb{R}^{n^i}$ is a tensor of order $i$, 
representing the coefficient tensor of $i$-order monomials.
$X_0$ is a constant $1$ by default.
The term 
$X_i = (x_{1\times n} \otimes x_{1\times n} \otimes \cdots \otimes x_{1\times n})$ 
represents the tensor product of the input vector repeated $i-1$ times.
The mark $\ast$ denotes the Hadamard (element-wise) product and $\otimes$ denotes the tensor product (outer product). 
For better understanding, Fig. \ref{tensor1} provides visual representations of tensor products. A special case of 
the mathematical expression (\ref{multipolynomial}) where $\boldsymbol{deg}(f) = 3$ is also shown in Fig. \ref{model1}.
$\boldsymbol{deg}()$  denotes the order (or degree) of a polynomial.

Similar to multiple linear regression (MLR), we fit the training samples by minimizing the squared error between the fitted values and 
the true values. Mathematically, this corresponds to solving the following optimization problem:
\begin{equation}
\begin{aligned} \label{lq}
&\min_{\boldsymbol{W_0} \in \mathbb{R}^1, \cdots, \boldsymbol{W_d}\in \mathbb{R}^{n^d}} 
\sum_{ p = 1}^{m}
( f\left((x_{1\times n})_p, \boldsymbol{W_0},\cdots \boldsymbol{W_d} \right) 
- \boldsymbol{y }_p)^2,
\end{aligned}
\end{equation}
where $(x_{1\times n})_p$ and $\boldsymbol{y}_p$ are the input feature vector and the label corresponding to $p$-th sample, respectively. 
There are $m$ samples for training in total. The goal is to find the optimal parameter tensors $\{ \boldsymbol{W_i}, i = 0, 1, 2, \cdots, d \}$.
To regularize the model and prevent overfitting, we can introduce $L_2$ regularization $\sum_{i = 0}^{d} \lambda_i || \boldsymbol{W_i} ||^2_F$ 
to penalize large parameters, where $\lambda_i$ controls the strength of regularization for the parameter tensor $\boldsymbol{W}_i$, and $\|\cdot\|_F$ 
represents the Frobenius norm.
For simplicity, we will not include $\sum_{i = 0}^{d} \lambda_i || \boldsymbol{W_i} ||^2_F$ in the following numerical experiments. 
However, differentiated regularization through $\lambda_i$ may play an important role in improving resistance to overfitting.

\subsection{Self-supervised Dimension Reduction}

In this subsection, we introduce our dimension reduction method aimed at addressing the Curse of Dimensionality.

Our approach relies on finding a low-rank linear transformation matrix $D_{n \times n}$ to reduce the requirement for the number of parameters 
while preserving the essential information of the data. Specifically, we solve the following optimization problem:
\begin{align}  \label{opt0}
\min_{D_{n \times n} \in \mathbb{R}^{n \times n}} 
Loss ( X_{m \times n} \cdot D_{n \times n} , X_{m \times n} ) ,\\
s.t. \quad \textbf{rank}(D_{n \times n}) =  k, k < n , \notag
\end{align}
where $X_{m \times n}$ denotes the matrix obtained by arranging $m$ sample feature vectors $x_{1 \times n}$ in rows.
The loss function measures the distance between the low-rank approximation of the input 
feature matrix and the original matrix. Utilizing the Frobenius norm, we express the loss as
$Loss = \left\lVert X_{m \times n} \cdot D_{n \times n} - X_{m \times n} \right\rVert_F^2$. 
Without loss of generality, we suppose that $\textbf{rank}(X_{m \times n}) = r > k$.
Since the rank of $D_{n \times n}$ is $k$, there exist matrices $L_{n \times k} $ and $ R_{k \times n}$ that satisfy the following decomposition:
\begin{align}
  D_{n \times n} = L_{n \times k} \cdot R_{k \times n}.
\end{align}
By decomposing $D_{n \times n}$ into $L_{n \times k} \cdot R_{k \times n}$ and treating 
$L_{n \times k}$ as a linear dimension reduction matrix applied to input feature vectors, we can effectively reduce the dimensionality 
of the input data while preserving important information for forecasting. 
The reduced data, denoted as $\tilde{X}_{m \times k}$, is obtained by applying the linear
transformation matrix $L_{n \times k}$ to the original data:
\begin{align}
  \tilde{X}_{m \times k} = X_{m \times n} \cdot L_{n \times k}.
\end{align}

This dimension reduction provides three main benefits:
\begin{enumerate}
  \item Reduce Computational Complexity: 
  Taking $d$-order terms as an example, when we denote the input feature matrix as
  $\widetilde{X}_{m \times k}$ (a single input sample feature vector is denoted as 
  $\widetilde{x}_{1 \times k}$), the $d$-order term sums to 
  $ \textbf{sum}((\boldsymbol{W_d})_{k \times k \cdots k}*\widetilde{X}_d) = \textbf{sum}((\boldsymbol{W_d})_{k \times k \cdots k}*(\widetilde{x}_{1 \times k} \otimes \widetilde{x}_{1 \times k} \otimes \ldots \otimes \widetilde{x}_{1 \times k}))$
  , where the number of parameters to optimize in the $d$-order parameter tensor
   $\boldsymbol{W_d}$ is reduced from $\mathcal{O}(n^d)$ to $\mathcal{O}(k^d)$.

  \item Preserve Model Fitting Capacity: The model maintains its ability to capture 
  complex relationships, as the dimension reduction is applied in a way that preserves 
  the core structure of the input features.
  For illustration, consider the quadratic term:

  \begin{align*}
    &\textbf{sum} \left((\boldsymbol{W_2})_{n \times n}* ( (x_{1 \times n} \cdot D_{n \times n}) \otimes (x_{1 \times n} \cdot D_{n \times n}) ) \right) \\
  = &(x_{1 \times n} \cdot D_{n \times n}) \cdot  (\boldsymbol{W_2} )_{n \times n} \cdot (
  D_{n \times n}^T \cdot x_{1 \times n}^T) \\
  = &(x_{1 \times n} \cdot L_{n \times k} ) \cdot \underbrace{R_{k \times n} \cdot  (\boldsymbol{W_2} )_{n \times n} \cdot 
  R_{k \times n}^T}_{(\boldsymbol{W_2'})_{k \times k}}  \cdot ( L_{n \times k}^T \cdot x_{1 \times n}^T) \\
  = &\widetilde{x}_{1 \times k} \cdot (\boldsymbol{W_2'})_{k \times k} \cdot \widetilde{x}_{1 \times k}^T \\
  = &\textbf{sum}((\boldsymbol{W_2'})_{k \times k}* ( \widetilde{x}_{1 \times k} \otimes \widetilde{x}_{1 \times k} ) )
  \end{align*}

  where $(\boldsymbol{W_2'})_{k \times k}$ and $(\boldsymbol{W_2})_{n \times n}$ denote $k$-dimensional $2$-order and $n$-dimensional $2$-order 
  optimizable parametric tensors (or matrices) respectively.
  For any $(\boldsymbol{W_2})_{n \times n}$, there exists a corresponding
  $(\boldsymbol{W_2'})_{k \times k} = R_{k \times n} \cdot  (\boldsymbol{W_2} )_{n \times n} \cdot R_{k \times n}^T$ 
  such that the above relationship holds for any $x_{1 \times n} \in \mathbb{R}^n$.
  This demonstrates that optimizing the polynomial model with $\widetilde{X}_{m \times k}$
  as the input feature matrix is equivalent to obtaining a low-rank approximation of $X_{m \times n}$ (i.e. $X_{m \times n}\cdot D_{n \times n}$)
  and then feeding it into the full-parameter polynomial model for optimization.
  When the polynomial order exceeds two, a similar process can be followed using tensor notation. 

  \item Overfitting Regularization: 
  After the embedding through the matrix $L_{n \times k}$, the number of parameters that can be freely adjusted in the model has been significantly reduced. 
  Appropriate $k$ can decrease the risk of overfitting.
\end{enumerate}

Additionally, we can apply different $k_i$ for different terms. As the order $i$ increases, the number of 
parameters in $\boldsymbol{W_i}$ increases exponentially. We can therefore choose a smaller $k_i$ when $i$ is larger
to keep the computation manageable. It is also important to emphasize that when obtaining the matrix $L_{n \times k}$, we can only use training set data and 
must not rely on future input features to avoid data leakage.

For convenience we denote the $i$-order optimizable parameter tensor $(\boldsymbol{W_i})_{k_i \times k_i \cdots \times k_i}$ as 
$(\boldsymbol{W_i})_{k_i}$ and denote 
$\underbrace{\widetilde{x}_{1 \times k_i} \otimes \widetilde{x}_{1 \times k_i} \cdots \otimes \widetilde{x}_{1 \times k_i}}_i = (x_{1 \times n} \cdot L_{n \times k_i}) \otimes (x_{1 \times n} \cdot L_{n \times k_i})\otimes \cdots (x_{1 \times n} \cdot L_{n \times k_i})$
as $\widetilde{X_i}$ $(i=1,2, \cdots d)$.
So our model after dimension reduction embedding can be represented as:
\begin{align} \label{multipolynomial2}
& \widetilde{f} \left(x_{1\times n}, \boldsymbol{W_0},(\boldsymbol{W_1})_{k_1},\cdots, (\boldsymbol{W_d})_{k_d}\right) \notag \\ =
&  \boldsymbol{W_0} 
+ \textbf{sum}\left( (\boldsymbol{W_1})_{k_1}*\widetilde{X_1}  \right) 
+\cdots  \textbf{sum} \left((\boldsymbol{W_d})_{k_d}*\widetilde{X_d}  \right), 
\end{align}
and the optimization problem can be expressed as:
\begin{align} \label{lq3}
  \min_{(\boldsymbol{W_i})_{k_i}\in \mathbb{R}^{(k_i)^{i}}} 
  \sum_{ p = 1}^{m}
  ( \widetilde{f} \left({(x_{1\times n})}_p, \boldsymbol{W_0},(\boldsymbol{W_1})_{k_1},\cdots, (\boldsymbol{W_d})_{k_d}\right) - \boldsymbol{y}_p )^2.
\end{align}

\begin{figure}[!t]
\centering
\includegraphics[width=6in]{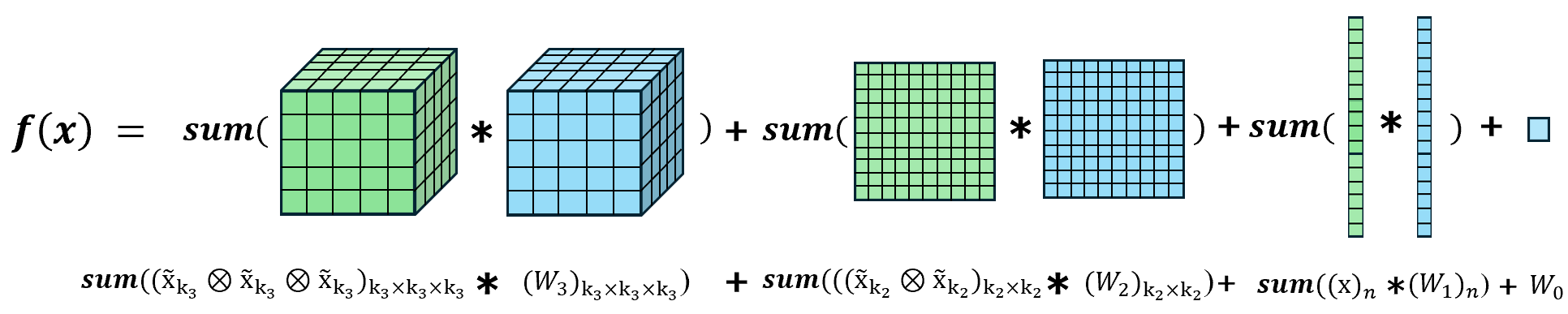}
\caption{An $n$-dimensional $3$-order polynomial with quadratic term embedding dimension $k_2$
and cubic term embedding dimension $k_3$.}
\label{model2}
\end{figure}

$\{k_i ,i=1,2,\cdots d \}$ is a set of hyperparameters for this model.
In practice, we can partition a validation set and perform grid search on it to obtain relatively suitable $k_i (i=1,2,\cdots )$. 
We can perform a grid search on the training set to find the optimal $\{k_i ,i=1,2,\cdots d \}$.
Typically, before conducting the search, we need to design an appropriate search range for $\{k_i ,i=1,2,\cdots d \}$ based on the specific context.
However, in some cases, additional hyperparameters beyond $\{k_i ,i=1,2,\cdots d \}$ may also need to be determined using the validation set. 
When the number of hyperparameters to be selected is too large, the curse of dimensionality can lead to prohibitively high computational 
costs for grid search. Therefore, to reduce computational demands, we may also fix appropriate $\{k_i ,i=1,2,\cdots d \}$ or some of them based on empirical knowledge, 
although this may lead to a slight decrease in the model's forecasting accuracy. 
For load forecasting tasks, the polynimoal model's performance is not highly sensitive to the choice of $k_i$, allowing us to make reasonable adjustments based on 
computational cost.
In the numerical experiments section of this paper, we adopt both above-mentioned two hyperparameter selection strategies depending on the specific scenario.

Up to this point, we have presented the full scope of our proposed polynimoal model.
Since the model is based on high-order polynomials with self-supervised dimensionality reduction, we refer to it as HOPS.
To provide a clearer presentation, we provide an illustration in Fig. \ref{model2}.

It is worth noting that if Z-score normalization is applied to the input data, the dimension reduction process described above becomes 
nearly equivalent to PCA in mathematical terms. 
In the case of load forecasting, the input variables are mostly categorical variables, so we generally do not use Z-score normalization.

\subsection{Properties and the Solution}

In this subsection, we provide some useful properties.

First, we highlight that the optimal solution to Problem (\ref{lq3}) exists, and the optimal solution set is either a single value or an affine space of 
$\mathbb{R}^{\sum_{i = 0}^d (k_i)^i}$. This is due to the fact that Problem (\ref{lq3}) can be reformulated as a large-scale linear 
regression problem. The specific details are provided in Theorem 3 of Appendix \uppercase\expandafter{\romannumeral2}.

Second, with regard to the optimization problem (\ref{opt0}), when the Frobenius norm is used as the loss function, an analytical solution can be derived 
by applying Singular Value Decomposition (SVD) and the Eckart-Young-Mirsky Theorem.
In particular, the solution requires performing Singular Value Decomposition on the input feature matrix $X_{m\times n}$, allowing us to decompose $X_{m\times n}$ 
in the following manner:
\begin{align}
  X_{m\times n} = U_{m\times m} \Sigma_{m \times n} V_{n \times n}^T.
\end{align}
Next, we take the first $k$ columns of $V_{n \times n}$ to obtain a required matrix $L_{n \times k}$. A detailed explanation of this process is provided in 
Theorem 2 of Appendix \uppercase\expandafter{\romannumeral1}.
When we choose a loss function other than the Frobenius norm, an analytical solution may no longer be available. In this case, the low-rank constraint can be 
regarded as a manifold, and the loss function can be optimized over this manifold. In the subsequent experiments, for simplicity, we use only the Frobenius 
norm as the loss function. However, the impact of different loss function choices remains an important consideration.

Additionally, there is an important point to consider here. Even though $D_{n \times n}$ is unique, $L_{n \times k}$ is not necessarily unique. 
So the mentioned-above $L_{n \times k}$ is merely one possible choice rather than the only one.
Any $L_{n \times k}$ can be transformed into a new matrix $L_{n \times k}'$ through an invertible linear transformation. This raises a natural concern: will 
the forecasting results change if we apply a different $L_{n \times k}$ for embedding?
Fortunately, we can prove that any $L_{n \times k} $ that satisfy the condition $D_{n\times n} = L_{n \times k} R_{k \times n}$, 
will be equivalent to our model. 
\begin{theorem}
Supposing that $\textbf{rank}(X_{m\times n}) > k_i,i=1,2,\cdots,d$ and for all $k_i$ (the embedding dimension of $i$-order terms), the solution to 
optimization problem (\ref{opt0}) is $D_{n \times n}^i$ 
$(\textbf{rank}(D_{n \times n}^i) = k_i, i = 1,2,\cdots d)$. For any 
$$
L_{n \times k_i} \in \mathbb{R}^{n \times k_i}, R_{k_i \times n} \in \mathbb{R}^{k_i \times n} , i = 1,2,\cdots d,
$$ and 
$$
L_{n \times k_i}' \in \mathbb{R}^{n \times k_i}, R_{k_i \times n}' \in \mathbb{R}^{k_i \times n} , i = 1,2,\cdots d,
$$
which satisfy
$$
D_{n \times n}^i = L_{n \times k_i} R_{k_i \times n} = L_{n \times k_i}' R_{k_i \times n}'  , i = 1,2,\cdots d,
$$
and there exists $i$ such that the following inequality holds
$$
L_{n \times k_i} \neq L_{n \times k_i}', R_{k_i \times n} \neq R_{k_i \times n}',
$$
the following conclusion can be drawn:
$L_{n \times k_i} $ and $ L_{n \times k_i}'$ ,$ i = 1,2,\cdots d $,
are equivalent in our model (\ref{multipolynomial2}). In other words, for any
$ \boldsymbol{W_0},(\boldsymbol{W_1})_{k_1},\cdots (\boldsymbol{W_d})_{k_d}$ ,
there exist 
$\boldsymbol{W_0'},(\boldsymbol{W_1'})_{k_1},\cdots (\boldsymbol{W_d'})_{k_d}$,
such that the following equation holds for any $\boldsymbol{x} \in \mathbb{R}^n$:
\begin{align*} 
& \widetilde{f}' \left(\boldsymbol{x}, \boldsymbol{W_0'},(\boldsymbol{W_1'})_{k_1},\cdots (\boldsymbol{W_d'})_{k_d} \right)
\\ & = \textbf{sum}(\boldsymbol{W_0'}*X_0') + \textbf{sum}((\boldsymbol{W_1'})_{k_1}*X_1') \cdots +
\textbf{sum} ((\boldsymbol{W_d'})_{k_d}*X_d')
\\ & = \textbf{sum}(\boldsymbol{W_0}*X_0) + \textbf{sum}((\boldsymbol{W_1})_{k_1}*X_1) \cdots +
\textbf{sum} ((\boldsymbol{W_d})_{k_d}*X_d),
\\ & = \widetilde{f} \left(\boldsymbol{x}, \boldsymbol{W_0},(\boldsymbol{W_1})_{k_1},\cdots (\boldsymbol{W_d})_{k_d} \right) ,
\end{align*}
where $X_i = \underbrace{\boldsymbol{x}L_{n \times k_i} \otimes \boldsymbol{x}L_{n \times k_i} \cdots \otimes \boldsymbol{x}L_{n \times k_i}}_i$
,$X_i' = \underbrace{\boldsymbol{x}L_{n \times k_i}' \otimes \boldsymbol{x}L_{n \times k_i}' \cdots \otimes \boldsymbol{x}L_{n \times k_i}' }_i, i = 1,2,\cdots d $
and $\boldsymbol{x} = x_{1\times n}$.
\end{theorem}
A detailed statement of and proof can be found in Theorem 4 of Appendix \uppercase\expandafter{\romannumeral3}.
A straightforward corollary of the above theorem is that for any two sets of distinct dimension reduction matrices 
$\{ L_{n\times {k_1}}, L_{n\times {k_2}}, \cdots L_{n\times {k_d}} \}$ and $\{ L'_{n\times {k_1}}, L'_{n\times {k_2}}, \cdots L'_{n\times {k_d}} \}$, 
the minimum value of the loss function obtained through optimization remains the same.
The above theorem essentially states that under different embedding matrices, the models can actually represent each other, which is what is referred to as "equivalent".

\section{A Fast Algorithm for Multivariate Higher-order Polynomial Model}

\subsection{Algorithm Overview}

In this section, we mainly describe the details about the fast algorithm applied to the embedded $n$-dimensional $d$-order polynomial model proposed 
in the previous section.

While we have already reduced the memory requirement from $\mathcal{O} (n^d)$ to $\mathcal{O} (k^d)$ by dimension reduction, the computational complexity 
still grows exponentially with $d$. Even with a relatively small $k$, the computational burden can remain significant.
Fortunately, the inherent mathematical structure of the polynomial model provides a way to address this challenge. 
By leveraging the tensor representation of $n$-dimensional $d$-order polynomials, we introduce a fast algorithm, as outlined in Algorithm \ref{alg:1}, 
to efficiently solve the Fitting Problem (\ref{lq3}). We name it as PolyCG (a modified FR-CG method for the polynomial model).
We have not made any fundamental modifications to FR-CG; instead, we have adapted it to the tensor representation form of the polynomial model 
and implemented a finite iteration truncation.
A simpler representation is to expand all terms represented by tensors and treat them as a hyper-dimensional input vector. 
In contrast, the advantage of the formulation in Algorithm \ref{alg:1} lies in separating the parameters of different orders, which allows for more convenient 
and targeted constraints on the parameters of different orders when needed. For example, as mentioned earlier, L2 regularization 
 can be applied to the parameters of different orders to prevent overfitting (i.e. $\sum_{i = 0}^{d} \lambda_i || (\boldsymbol{W_i})_{k_i} ||^2_F$).
However, in order to avoid making the paper too lengthy, we will not discuss the case of applying targeted penalization.

For simplicity, in this subsection we denote $\widetilde{f}^j\left(\boldsymbol{x}\right)$ and $\widetilde{f}^j_{\boldsymbol{R}}\left(\boldsymbol{x}\right)$ as a shorthand for
$\widetilde{f}\left(\boldsymbol{x},\boldsymbol{W_0}^j,(\boldsymbol{W_1})_{k_1}^j,\cdots, (\boldsymbol{W_d})_{k_d}^j \right)$ and
$\widetilde{f}\left(\boldsymbol{x},\boldsymbol{R}_0^j,\boldsymbol{R}_1^j,\cdots, \boldsymbol{R}_d^j \right)$ respectively. 
We also represent 
$(\underbrace{x_{1\times n}L_{n \times k_i} \otimes x_{1\times n}L_{n \times k_i} \otimes \cdots x_{1\times n}L_{n \times k_i}}_i  )$
as $ \widetilde{X}_i(x) $.
The superscript $j$ in the context all represents the $j$-th iteration.
Additionally, $\alpha^j$ is computed as follows:
\begin{align}
\alpha^j = \frac{\sum_{i=0}^{d}
\left\Vert \boldsymbol{R}_i^j \right\Vert _F^2}
{2 \left( \sum_{ p = 1}^{m} (\widetilde{f}^j_{\boldsymbol{R}} \left( \boldsymbol{x}_p \right) 
- \boldsymbol{0})^2 \right)},
\end{align}
and $\boldsymbol{R}_i^j$ is computed by:
\begin{align}
\boldsymbol{R}_i^j = 
\frac{\partial \left( \sum_{ p = 1}^{m} (\widetilde{f}^j \left( \boldsymbol{x}_p \right) - \boldsymbol{y}_p)^2 \right)}
{\partial (\boldsymbol{W}_i)^j_{k_i} }
&= \sum_{ p = 1}^{m}
2(\widetilde{f}^j \left(\boldsymbol{x}_p \right) - \boldsymbol{y}_p ) \widetilde{X}_{i}(\boldsymbol{x}_p ).
\end{align}

Owing to the favorable mathematical properties of $n$-dimensional $d$-order polynomials and the tensor-based formulation defined in equation
(\ref{multipolynomial}), any derivative process can be analytically expressed as a simple combination of additions and multiplications between 
tensors or tensors and scalars. Similar to the optimization advantages observed in deep neural networks, this characteristic is well-suited for GPU 
parallel computing, as GPUs excel at performing simple but large-scale matrix and tensor operations. 
Consequently, these numerical computations can be efficiently handled by PyTorch, Nvidia GPUs, and CUDA with minimal effort.

\begin{algorithm}
	\renewcommand{\algorithmicrequire}{\textbf{INPUT:}}
	\renewcommand{\algorithmicensure}{\textbf{OUTPUT:}}
	\caption{PolyCG} 
	\label{alg:1}
	\begin{algorithmic}[1]
  \REQUIRE Training Feature Set $\boldsymbol{X}$; Training Label Set $\boldsymbol{Y}$;
    Initial Values of the Parameter Tensors $(\boldsymbol{W}_i)_{k_i}^0 , i = 0,1,2,\cdots d$; 
    Embedding Dimensions For Different Order Terms $\{k_i,i=1.2,\cdots d\}$;
    Maximum Iteration $K=1000$; the Tolerance $\epsilon = 1e-7$.
		\ENSURE  Values of the parameter tensors $(\boldsymbol{W}_i)_{k_i}^j , i = 0,1,2,\cdots d$;

		\STATE Calculate
    $\boldsymbol{R}^0_i, i = 0,1,2,\cdots d$;

    \STATE Let
    $\boldsymbol{P}^0_i = - \boldsymbol{R}^0_i, i = 0,1,2,\cdots d$;

		\FOR {$j = 0$ to $K-1 $}
		\STATE Calculate $\alpha^j$;

    \STATE $(\boldsymbol{W}_i)_{k_i}^{ j + 1 }
    = (\boldsymbol{W}_i)_{k_i}^{ j} +\alpha^j \boldsymbol{P}^j_i, i = 0, 1,2,\cdots d$;

    \STATE Obtain $\widetilde{f}^{j+1}$ by $(\boldsymbol{W}_i)_{k_i}^{ j + 1 }, i = 0, 1,2,\cdots d$;

    \STATE If $\frac{| \sum_{ p = 1}^{m} (\widetilde{f}^{j+1} \left( \boldsymbol{x}_p \right) - \boldsymbol{y}_p)^2 - 
    \sum_{ p = 1}^{m} (\widetilde{f}^j \left( \boldsymbol{x}_p \right) - \boldsymbol{y}_p)^2|}
    {\sum_{ p = 1}^{m} (\widetilde{f}^j \left( \boldsymbol{x}_p \right) - \boldsymbol{y}_p)^2} 
    \leq \epsilon$,
     terminate the loop. \label{terminate}

		\STATE Calculate $\boldsymbol{R}_i^{j+1}, i = 0,1,2,\cdots d$;

    \STATE $\beta^j = \frac{\sum_{i=0}^{d}
        \left\Vert \boldsymbol{R}_i^{j+1} \right\Vert_F^2}
        {\sum_{i=0}^{d}
        \left\Vert \boldsymbol{R}_i^{j} \right\Vert_F^2} $;

    \STATE  Calculate $\boldsymbol{P}_i^{j+1} = 
        -\boldsymbol{R}_i^{j+1} + \beta^j \boldsymbol{P}_i^{j}, i = 0,1,2,\cdots d$;
		\ENDFOR
		\STATE \textbf{Return} Values of the parameter tensors 
    $(\boldsymbol{W}_i)_{k_i}^j , i = 0,1,2,\cdots d$.
	\end{algorithmic}  
\end{algorithm}

\subsection{A Note on Algorithm \ref{alg:1}}

Essentially, the solution of the proposed model can be interpreted as a large-scale linear regression problem.
This provides a simple and intuitive way to conceptualize the polynomial model. 
By extending the feature space geometrically, the model can fit the dataset in a much higher-dimensional space. 
However, this approach is often computationally impractical. 
Super-high-dimensional linear regression typically involves the inversion and multiplication of extremely large matrices, 
which are both computationally expensive and memory-intensive, especially when $\sum_{i = 0}^d (k_i)^i$ is too large.
Given  the necessity to solve large-scale linear equations or optimization problems under limited computational and 
storage resources, the CG method presents itself as a viable solution \cite{saad2003iterative}. 
This method has been well-studied and proven effective in such contexts \cite{golub2013matrix}.
Therefore, building on the core principles of CG, we modify the Fletche Reeves CG algorithm (FR-CG)
\cite{fletcher1964function} to provide a practical and efficient approach, as presented in Algorithm \ref{alg:1}, specifically adapted to solve the 
optimization problem (\ref{lq3}). 

It is worth noting that in practice, the number of iterations required is often significantly 
less than $\sum_{i = 0}^d (k_i)^i$, which is required for an exact solution theoretically. 
Due to round-off error, when the problem scale is too large, the algorithm only obtains an approximate numerical solution.

\section{Numerical Experiment}
\subsection{Datasets and Metrics}

\subsubsection{Datasets}
The datasets utilized for our numerical experiments are from the ISO New England (ISONE). ISO New England Inc. is an independent, non-profit regional transmission organization (RTO).
These datasets were also used in Global Energy Forecasting Competition 2017. 
These datasets are chosen due to the extensive body of research conducted on it, facilitating easier comparisons with existing methods.
Specifically, the datasets we used consists of 10 hourly load datasets, covering eight load zones: Maine (ME), New Hampshire (NH), Vermont (VT), 
Connecticut (CT), Rhode Island (RI), Southeastern Massachusetts (SEMASS), Western Central Massachusetts (WCMASS), 
and Northeastern Massachusetts (NEMASS). 
Additionally, it includes the combined load for the three Massachusetts regions (MASS) and the total combined load for all regions (TOTAL). 
These datasets also provides hourly drybulb temperature and dew point temperature data. 
Relative humidity can be calculated from drybulb temperature and dew point temperature by the Tetens equation:
\begin{align}
 {RH}_t = \frac{e^{17.27 \times \frac{{DEW}_t}{{DEW}_t + 237.3}}}{e^{17.27 \times \frac{T_t}{T_t+237.3}}} \times 100\%,
\end{align}
where ${DEW}_t$ and $T$ represent the dew point temperature and the temperature (dry bulb temperature) at time $t$, respectively.
Since the dew point temperature and the temperature data provided by ISONE is in Fahrenheit ($F^{\circ}$), and those in the Tetens equation is in Celsius ($C^{\circ}$), 
we first convert the dew point temperature and the temperature to Celsius before calculating the relative humidity ${RH}_t$. 
In the subsequent sections, when temperature is used as a variable, we retain the original temperature unit ($F^{\circ}$).

In the following, we pick three years (2012-2014) of hourly data for training and one year (2015) for testing. 
However, when comparing with the Recency model \cite{wang2016electric}, we select two years (2013-2014) of training data to remain consistent with the original literature. 

\subsubsection{Evaluation Metrics}
Although various evaluation metrics are employed for load forecasting, Mean Absolute Percentage Error (MAPE) is most commonly used due to 
its transparency and simplicity \cite{hong2013long, luo2023robust, xie2017variable}. 
The MAPE is defined as:
\begin{align*}
MAPE = \frac{1}{N} \sum_{t=1}^{N} \left\vert \frac{y_t - \widehat{y}_t }{y_t} \right\vert \times 100\%,
\end{align*}  
where $y_t$ and $\widehat{y}_t$ are the actual load and the predicted load at time $t$, respectively.
$N$ denotes the number of samples.

To diversify the evaluation metrics, we also utilize another widely adopted metric, 
Mean Squared Error (MSE), which is also commonly used in load forecasting studies, 
such as in \cite{luo2023robust}. The MSE is defined as:
\begin{align*}
\text{MSE} = \frac{1}{N} \sum_{t=1}^{N} \left({y_t - \widehat{y}_t }\right)^2,
\end{align*}  
where $y_t$, $ \widehat{y}_t $ and $ N $ are the same as above.

\subsection{Implementation Details}
All numerical experiments are conducted on a desktop equipped with a 12th Gen Intel(R) i5-12600KF CPU, 32.0 GB usable RAM, H800 GPU 
with 80GB RAM and Microsoft Windows 11 Enterprise. We implement our algorithm through Python 3.11.5, CUDA 12.6 and Jupyter Notebook 6.5.4.

For hyperparameter settings, we set the maximum number of iterations to be 1000 and the tolerance $\epsilon = 1e-7$.
Similar to \cite{luo2023robust}, we normalize all input variables by Min-Max normalization. 
Since the primary term is generally less prone to overfitting, we set $k_1$ equal to the sample feature dimension $n$.
Given the relatively straightforward relationship between load and features, we limit the max polynomial order $d=3$, rather than opting for a larger order.
The optimal embedding dimensions $k_2$ and $k_3$ are chosen by a validation set (the data of 2014). We retrain the model on the whole training set (2012-2014) after getting $k_2$ and $k_3$ on the validation set.

The trend item $Trend_t$ is not treated as a typical feature variable. 
Given its particular nature, we include it in the primary term but exclude it from the higher-order terms and then remove it before performing dimension reduction.
So the highest optional embedding dimension for the higher-order terms is set to be $n-1$ ($1$ less than the input feature dimension). 
Regarding the selection of $k_2$ and $k_3$, when the input variables are fixed, we primarily employ the method of partitioning a validation set and conducting 
a grid search to determine the optimal parameter.
We first train the model on data from 2012-2013 and then use data from 2014 as the validation set to determine the optimal hyperparameter 
$k_2$ and $k_3$. Subsequently, we fix the selected optimal $k_2$ and $k_3$ and retrain the model using data from 2012-2014 before forecasting.
When we perform a grid search to select the optimal embedding dimensions $k_2$ and $k_3$, we set the quadratic term embedding dimension range to be 
$[12, 24, 36, 48, 60, 72, 84, 96 ,108]$ and the cubic term embedding dimension range to be $[0, 1, 5, 9, 13, 17, 21]$ for data with a feature dimension of 289.
For other input data, we set the quadratic term embedding dimension range to be $[20, 28, 36, 44, 52, 60, 68, 76, 84]$ (if greater than $n-1$, replace it with $n-1$)
and the cubic term embedding dimension range to be $[0, 1, 5, 9, 13, 17, 21]$.
For practicality, when using a variable selection method similar to the Recency model (i.e. ReHOPS), we fix $k_2 = 60$ and $k_3 = 9$ for all 10 datasets.
This variable selection method also requires traversal on the validation set to choose the appropriate input variables.
Performing a grid search over all three hyperparameters simultaneously would be much-too time-consuming.

\subsection{The Competitive Comparison Models}

Hundreds of models have been developed for load forecasting \cite{hong2010short}. 
However, as noted in \cite{hong2013long, luo2023robust}, one of the most frequently cited is constructed in the following manner:
\begin{align*} 
y_t= & r_0 + r_1 Trend_t + 
r_2 H_t + r_3 W_t+r_4 M_t \\ & + r_5 H_t W_t + f(T_t),
\end{align*}
where $y_t$ denotes the load value that needs to be predicted at time $t$, 
$H_t$ is dummy variables (a vector) corresponding to 24 hours per day.
$W_t$ is dummy variables (a vector) indicating 7 days of a week.
$M_t$ is dummy variables (a vector) indicating 12 months of a year. 
$T_t$ denotes the concurrent temperature.
$Trend_t$ is a variable of the increasing integers representing a linear trend and 
  \begin{align*} 
f \left( T \right)= & \beta_1 T + \beta_2 \left(T\right)^2 + \beta_3 \left(T\right)^3 + \beta_4 T_t H_t + \beta_5 \left( T \right)^2
H_t + \\ &  \beta_6 \left(T\right)^3 + \beta_7 T M_t + \beta_8 \left(T\right)^2 M_t
+ \beta_9 \left(T\right)^3 M_t.
\end{align*}
In \cite{luo2023robust}, the authors did not include $\left(T_{t} \right)^k$ as variables in their regression model to avoid perfect 
multicollinearity. Similarly, $H_t$ and $W_t$ were excluded for the same reason.
Following the approach outlined in \cite{luo2023robust}, we combine the variables to obtain 289-dimensional input variables for MLR. 
As a widely recognized and effective method for load forecasting, this feature-based MLR model has been frequently cited and utilized as a benchmark 
in GEFCom2012, GEFCom2014, and GEFCom2017. 
We refer to this vanilla model as $G_{1}$.

In \cite{xie2016relative}, the authors highlighted the significant role of humidity in load forecasting and recommended a model as follows:
\begin{align*} 
  y_t= g(RH_t) + G,
\end{align*}
where $G$ represents a base model depending upon temperature variables, and 
\begin{align*} 
  g(RH_t) =& \gamma_1 RHS_t + \gamma_2 RHS_t^2 + \gamma_3 T_t \times RHS_t
  + \gamma_4 T_t^2 \times RHS_t \\ & + \gamma_5 T_t \times RHS_t^2 + 
  \gamma_6 T_t^2 \times RHS_t^2  + \gamma_7 H_t \times RHS_t \\ & + 
  \gamma_8 H_t \times RHS_t^2.
\end{align*} 
$H_t$ and $T_t$ are the same as before.
$RH$ denotes relative humidity (\%). 
$S_t$ denotes a dummy variable indicating Jun.-Sep. of a year.
$RHS_t$ denotes cross effect of relative humidity and summer.
$RHS_t^2$ denotes cross effect of relative humidity square and summer.
All the above symbols are consistent with \cite{xie2016relative}.
We take $G = G_{1}$ and denote $ y_t= g(RH_t) + G_1 $ as $H_1$.

Additionally, following the approach in \cite{wang2016electric}, we incorporate lagged temperature variables and obtain an improved 
model that only depends on the temperature and date:
  \begin{align*} 
  y_t= & r_0 + r_1 Trend_t + 
  r_2 H_t + r_3 W_t+r_4 M_t \\ & + r_5 H_t W_t + \sum_{h=0}^{3} f(T_{t-h}),
  \end{align*}

where $H_t, W_t, M_t$ are the same as above. $T_{t-h}$ denotes the temperature of the previous $h$-th hour.
As before, $\left(T_{t-h} \right)^k$, $H_t$ and $W_t$ are excluded to avoid perfect multicollinearity.
This results in a total of 613 variables and we refer to it as $G_{2}$.
By incorporating both humidity information and lagged temperature variables, we present this second improved model for comparison:
\begin{align*} 
  y_t= g(RH_t) + G_2.
\end{align*}
We denote this model as $H_2$. 

In conclusion, these two improved models, $H_1$ and $H_2$, contain $343$ and $667$ variables, respectively.
To present different models more clearly, we have listed the input variables of various models in Table \ref{variable}.

\begin{table}[htbp]
  \centering
  \caption{Feature Selection for Various Models}
  \renewcommand\arraystretch{1.5}
  \begin{tabular}{|c|c|c|}
  \hline			
  MODEL                       & \text{Feature}                                    & \text{Dimension} \\
  \hline		
  \multirow{3}*{HOPS$289$}	& $Trend_t$, $H_t$, $W_t$, $M_t$, $T_t$, $(T_t)^2$, & \multirow{3}*{$289$}\\
                              & $(T_t)^3$, $T_tH_t$, $(T_t)^2H_t$, $(T_t)^3H_t$   &                     \\
                              & $W_tH_t$, $T_tM_t$, $(T_t)^2M_t$, $(T_t)^3M_t$    &                     \\
  \hline		
  \multirow{2}*{HOPS$47$}	  & $Trend_t$, $H_t$, $W_t$, $M_t$, $T_t$,            & \multirow{2}*{$47$} \\
                              & $(T_t)^2$,$(T_t)^3$                               &                     \\
  \hline		
  \multirow{2}*{HOPS$50$}	  & $Trend_t$, $H_t$, $W_t$, $M_t$, $T_t$, $(T_t)^2$, & \multirow{2}*{$50$} \\
                              &  $(T_t)^3$ ,$ RH_t$, $(RH_t)^2$, $(RH_t)^3$       &                     \\
  \hline		
  \multirow{3}*{HOPS$59$}	  & $Trend_t$, $H_t$, $W_t$, $M_t$, $T_t$, $(T_t)^2$, & \multirow{3}*{$59$} \\
                              & $(T_t)^3$ ,$ RH_t$, $(RH_t)^2$, $(RH_t)^3$,       &                     \\
                              & $T_{t-i}$, $(T_{t-i})^2$, $(T_{t-i})^3$, $i=1,2,3$&                     \\
  \hline			
  \multirow{3}*{ $G_1$}	      & $Trend_t$, $H_t$, $W_t$, $M_t$, $T_t$, $(T_t)^2$, & \multirow{3}*{$289$}\\
                              & $(T_t)^3$, $T_tH_t$, $(T_t)^2H_t$, $(T_t)^3H_t$   &                     \\
                              & $W_tH_t$, $T_tM_t$, $(T_t)^2M_t$, $(T_t)^3M_t$    &                     \\
  \hline		
  \multirow{6}*{ $H_1$}	      & $Trend_t$, $H_t$, $W_t$, $M_t$, $T_t$, $(T_t)^2$, & \multirow{6}*{$343$}\\
                              & $(T_t)^3$, $T_tH_t$, $(T_t)^2H_t$, $(T_t)^3H_t$   &                     \\
                              & $RHS_t$, $RHS_t^2$, $T_t \times RHS_t$,           &                     \\
                              & $T_t^2 \times RHS_t$, $T_t \times RHS_t^2$,       &                     \\
                              & $T_t^2 \times RHS_t^2$, $H_t \times RHS_t$,       &                     \\
                              & $H_t \times RHS_t^2$.                             &                     \\
  \hline	
  \multirow{10}*{ $H_2$}	    & $Trend_t$, $H_t$, $W_t$, $M_t$, $T_t$, $(T_t)^2$, &\multirow{10}*{$613$}\\
                              & $(T_t)^3$, $T_tH_t$, $(T_t)^2H_t$, $(T_t)^3H_t$   &                     \\
                              & $W_tH_t$, $T_tM_t$, $(T_t)^2M_t$, $(T_t)^3M_t$    &                     \\
                              & $T_{t-i}$, $(T_{t-i})^2$, $(T_{t-i})^3$,          &                     \\
                              & $T_{t-i}H_t$, $(T_{t-i})^2H_t$, $(T_{t-i})^3H_t$, &                     \\
                              & $T_{t-i}M_t$, $(T_{t-i})^2M_t$, $(T_{t-i})^3M_t$, &                     \\
                              & $RHS_t$, $RHS_t^2$, $T_t \times RHS_t$,           &                     \\
                              & $T_t^2 \times RHS_t$, $T_t \times RHS_t^2$,       &                     \\
                              & $T_t^2 \times RHS_t^2$, $H_t \times RHS_t$,       &                     \\
                              & $H_t \times RHS_t^2$, $i=1,2,3$.                  &                     \\
  \hline
\end{tabular}
\label{variable}
\end{table}

\subsection{Overall Performance with the Same or Simpler Inputs}

First, to demonstrate that the improved performance of our proposed method is due to the method itself rather than changes in variables, we 
utilize the same variables from the vanilla model $G_{1}$ as the underlying variables for our proposed model, which is named as 
HOPS289 (the HOPS Model with 289 variables) for distinction.
The experimental results are presented in Table \ref{table1}. 
It is evident that our model outperforms the vanilla model $G_{1}$ in terms of load forecasting accuracy across all regions, 
regardless of whether MAPE or MSE is used as the evaluation metric.

In \cite{hong2013long, luo2023robust}, the authors selected and constructed 289 variables based on their practical significance to get the vanilla model $G_{1}$, 
whose high accuracy is largely attributed to the complexity of the variable construction process. However, this process can be somewhat tedious. 
Our goal is to achieve better prediction performance with simpler input variables, which will be more beneficial for including more relevant variables and solving the model.
Fortunately, we have found that our model can achieve comparable forecasting performance without the necessity for such intricate variable construction.

Specifically, rather than constructing interaction terms, we utilize the same underlying input information (temperature and date) but with fewer 
variables compared to the vanilla model $G_{1}$. The selected input variables are 
$\{ Trend_t$, $ H_t$, $ W_t$, $ M_t$, $ T_t$, $ (T_t)^2 $, $ (T_t)^3 \}$, 47 variables in total. We refer to this model as HOPS47. 
To assess the impact of complex versus simple input variables on our model's performance, we compare the forecasting results utilizing
$47$-dimensional input variables against those utilizing $289$-dimensional input variables. 
The experimental results and the comparison are also presented in Table \ref{table1}.

The experimental results indicate that the two sets of input variables make little difference in our model's performance. The accuracy of both 
HOPS47 and HOPS289 is nearly identical, and both outperform the vanilla model $G_{1}$.
Although, on average, HOPS289 performs slightly better than HOPS47, the variable construction of HOPS47 is simpler and easier. Moreover, the computation time for HOPS47 is significantly lower than that of HOPS289, as HOPS289 typically requires a broader 
search range for the embedding dimension $k_2$ and $k_3$, which means that it will take us more time to train the model.
It takes 19 minutes 24 seconds to finish all the numerical experiments for HOPS47 while 66 minutes 9 seconds for HOPS289.
We recommend using the variable input method of HOPS47.

\begin{table}[htbp]
  \centering
  \caption{Comparisons Between  $G_{1}$ and HOPS289, HOPS47. }
    \begin{tabular}{m{0.12\textwidth}<{\raggedright} | m{0.07\textwidth}<{\raggedleft} m{0.07\textwidth}<{\raggedleft} 
      m{0.07\textwidth}<{\raggedleft} | m{0.07\textwidth}<{\raggedleft} m{0.07\textwidth}<{\raggedleft} m{0.07\textwidth}<{\raggedleft} }
    \toprule
     & \multicolumn{3}{c|}{MAPE(\%)} & \multicolumn{3}{c}{MSE (×10$^3$)} \\
    \cmidrule{2-3} \cmidrule{4-7} 
      & \multicolumn{2}{c}{HOPS} & \multirow{2}{*}{\textbf{$G_{1}$}} & \multicolumn{2}{c}{HOPS} & \multirow{2}{*}{\textbf{$G_{1}$ }} \\
    \cmidrule{2-3} \cmidrule{5-6} 
    Zone & 289dim & 47dim &  & 289dim & 47dim & \\
    \midrule
    CT             & \textbf{3.99}  & \underline{4.03} & 4.18  & \textbf{38.68}  & \underline{38.76} & 42.52  \\
    MASS           & \textbf{3.60}  & \underline{3.64} & 3.81  & \textbf{109.80} &\underline{110.51} & 122.42 \\
    ME             & \textbf{4.38}  & \textbf{4.38}    & 4.50  & \textbf{4.96}   &\underline{4.98}  & 5.25 \\
    NEMASS         & \textbf{3.66}  & \textbf{3.66}    & 3.90  & \textbf{25.36}  &\underline{25.48} & 28.53 \\
    NH             & \underline{3.52}  & \textbf{3.48} & 3.73  & \underline{4.21}&\textbf{4.12}  & 4.75 \\
    RI             & \textbf{3.64}  & \underline{3.72} & 3.86  & \textbf{2.40}   &\underline{2.46}  & 2.78 \\
    SEMASS         & \textbf{4.31}  & \underline{4.40} & 4.64  & \textbf{10.41}  &\underline{10.78} & 11.98 \\
    VT             & \textbf{4.08}  & \textbf{4.08} & 4.29  & \underline{1.21}&\textbf{1.18}  & 1.29 \\
    WCMASS         & \textbf{4.28}  & \underline{4.30} & 4.41  & \underline{13.04}&\textbf{12.86} & 13.78 \\
    \midrule
    AVERAGE & \textbf{3.94} & \underline{3.97} & 4.15 & \textbf{23.34} & \underline{23.46} & 25.92 \\
    \midrule
    TOTAL & \textbf{3.37} & \underline{3.41} & 3.54 & \textbf{454.62} & \underline{457.13} & 501.55 \\
    \bottomrule
    \end{tabular}
  \label{table1}
\end{table}

\subsection{Overall Performance with Other Advanced Models}

Currently, many load forecasting models primarily improve forecasting accuracy by selecting more relevant variables, such as incorporating humidity 
information \cite{xie2016relative} or leveraging recency effects \cite{wang2016electric}.
However, these models merely increase the input variables without considering improving performance by altering the model's fitting function. 
Below, we compare the advantages of our model over these advanced models under the same input information.

Similar to HOPS47, we do not construct interaction terms and use the same temperature, relative humidity, and date information as in $H_1$ and $H_2$. 
Specifically, for $H_1$, the variables include
\{ $Trend_t$, $H_t$, $W_t$, $M_t$, $T_t$, $(T_t)^2$, $(T_t)^3$, $ RH_t$, $(RH_t)^2$, $(RH_t)^3$ \},
resulting in a total of 50 variables.
For $H_2$, the variables include
\{$Trend_t$, $ H_t$, $W_t$, $ M_t$, $T_t$, $(T_t)^2$, $(T_t)^3 $, $ RH_t$, $(RH_t)^2$ $(RH_t)^3 $, $ T_{t-i}$, $(T_{t-i})^2$, $(T_{t-i})^3$, $i = 1,2,3$ \},
totaling 59 variables.
We refer to these models as HOPS50 and HOPS59, respectively. 
$H_{1}$ and HOPS50 share the same input information, although there are different input variables for these two models.
The relationships between $G_{1}$ and HOPS47, $H_{2}$ and HOPS59 are similar to $H_{1}$ and HOPS50.
The experimental results and the comparison are presented in Table \ref{table2}.
It can be observed that our model consistently demonstrates a stable advantage when the input information is the same. 
Moreover, as the input information increases, our model also shows steady improvement.

\begin{table*}[htbp]
  \centering
  \caption{Comparisons Between the Vanilla Model $G_{1}$ and HOPS47, the improved model $H_{1}$ and HOPS50, the improved model $H_{2}$ and HOPS59. 
  $G_{1}$ and HOPS47,  $H_{1}$ and HOPS50, $H_{2}$ and HOPS59 share the same input information respectively.}
    \begin{tabular}{m{0.08\textwidth}<{\raggedright} | m{0.045\textwidth}<{\raggedleft} m{0.045\textwidth}<{\raggedleft}  | m{0.045\textwidth}<{\raggedleft} 
      m{0.045\textwidth}<{\raggedleft}  | m{0.045\textwidth}<{\raggedleft} m{0.045\textwidth}<{\raggedleft}  |
      m{0.055\textwidth}<{\raggedleft}  m{0.055\textwidth}<{\raggedleft} |m{0.052\textwidth}<{\raggedleft} m{0.052\textwidth}<{\raggedleft} |
      m{0.052\textwidth}<{\raggedleft} m{0.052\textwidth}<{\raggedleft} }
      \toprule
      & \multicolumn{6}{c|}{MAPE(\%)} & \multicolumn{6}{c}{MSE (×10$^3$)} \\
      \cmidrule{2-4} \cmidrule{5-13}
       Zone & HOPS47 & \textbf{$G_{1}$} & HOPS50 & \textbf{$H_{1}$} & HOPS59 & \textbf{$H_{2}$} 
       & HOPS47 & \textbf{$G_{1}$} & HOPS50 & \textbf{$H_{1}$} & HOPS59 & \textbf{$H_{2}$} \\
      \midrule
    CT             & \textbf{4.03} & 4.18  & \textbf{3.64} & 4.02  & \textbf{3.37} & 3.75 & \textbf{38.76} & 42.52 & \textbf{30.23} & 37.08  & \textbf{25.99} & 32.14  \\
    MASS           & \textbf{3.64} & 3.81  & \textbf{3,14} & 3.53  & \textbf{2.96} & 3.36 & \textbf{110.51} & 122.42 & \textbf{80.43} & 99.82  & \textbf{70.51} & 89.38  \\
    ME             & \textbf{4.38} & 4.50  & \textbf{4.33} & 4.41  & \textbf{4.28} & 4.36 & \textbf{4.98}  & 5.25 & \textbf{4.82} & 5.06  & \textbf{4.67} & 4.92 \\
    NEMASS         & \textbf{3.66} & 3.90  & \textbf{3.11} & 3.49  & \textbf{2.93} & 3.33 & \textbf{25.48} & 28.53 & \textbf{16.94} & 21.06  & \textbf{14.75} & 18.75 \\
    NH             & \textbf{3.48} & 3.73  & \textbf{3.15} & 3.49  & \textbf{2.89} & 3.24 & \textbf{4.12}  & 4.75 & \textbf{3.23} & 3.98  & \textbf{2.69} & 3.39  \\
    RI             & \textbf{3.72} & 3.86  & \textbf{3.47} & 3.68  & \textbf{3.24} & 3.47 & \textbf{2.46}  & 2.78 & \textbf{2.01} & 2.23  & \textbf{1.72} & 1.95  \\
    SEMASS         & \textbf{4.40} & 4.64  & \textbf{3.92} & 4.37  & \textbf{3.65} & 4.15 & \textbf{10.78} & 11.98 & \textbf{8.22} & 9.48  & \textbf{6.84} & 8.39  \\
    VT             & \textbf{4.08} & 4.29  & \textbf{3.60} & 4.06  & \textbf{3.47} & 3.89 & \textbf{1.18}  & 1.29 & \textbf{0.90} & 1.13  & \textbf{0.85} & 1.04  \\
    WCMASS         & \textbf{4.30} & 4.41  & \textbf{3.74} & 4.13  & \textbf{3.58} & 3.98 & \textbf{12.86} & 13.78 & \textbf{9.88} & 11.91  & \textbf{9.01} & 11.04 \\
    \midrule
    AVERAGE & \textbf{3.97} & 4.15  & \textbf{3.57} & 3.91  & \textbf{3.37} & 3.73 & \textbf{23.46} & 25.92 & \textbf{17.60} & 21.30  & \textbf{15.22} & 19.00  \\
    \midrule
    TOTAL & \textbf{3.41} & 3.54  & \textbf{2.95} & 3.32  & \textbf{2.75} & 3.09 & \textbf{457.13} & 501.55 & \textbf{323.86} & 412.89  & \textbf{280.41} & 359.24  \\
    \bottomrule
    \end{tabular}
  \label{table2}
\end{table*}

\begin{figure}[!t]
\centering
\includegraphics[width=6.5in]{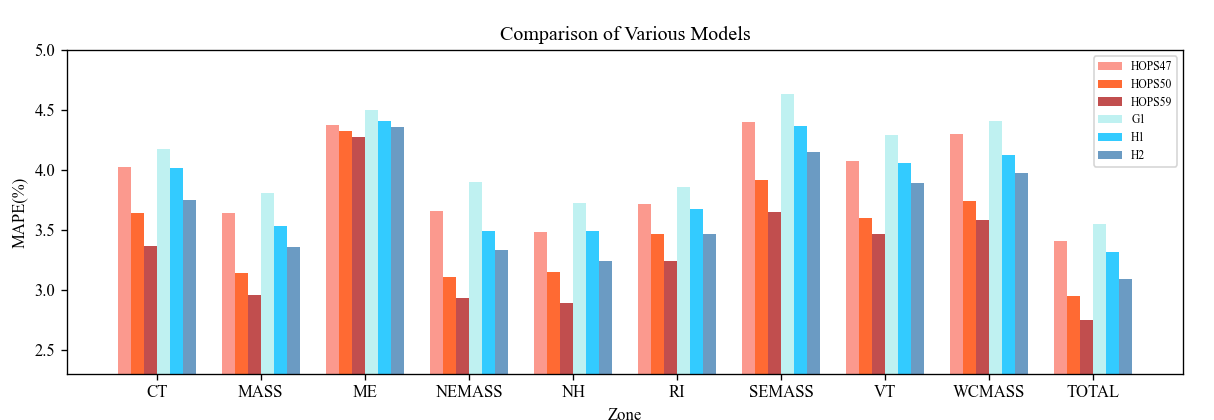}
\caption{The performance comparision of $G_1$, $H_1$, $H_2$, HOPS47, HOPS50 and HOPS59. $G_1$ and HOPS47, $H_1$ and HOPS50, $H_2$ and HOPS59 essentially 
share the same information, respectively. The forecasting accuracy of these models generally improves as the number of variables increases. However, when the input 
information keeps the same, our model consistently outperforms the existing models $G_1$, $H_1$ and $H_2$, respectively.}
\label{zone}
\end{figure}

To further clarify the results, we plot the performances of $G_1$, $H_1$, $H_2$, HOPS47, HOPS50 and HOPS59 as Fig. \ref{zone}.
$G_1$ and HOPS47, $H_1$ and HOPS50, $H_2$ and HOPS59 essentially share the same information, respectively.
As shown, the forecasting accuracy of the models generally improves as the number of variables increases. However, when the input 
information is the same, our model consistently outperforms the existing models $G_1$, $H_1$ and $H_2$.

\subsection{Extension Experiments Compared with the Recency Model}

Our polynomial model operates independently of variable selection, which implies that it can be integrated with other existing variable selection based models to further 
enhance forecasting accuracy. We take the Recency model as an example for further discussion.

As an enhanced variable selection based model for load forecasting, the Recency model \cite{wang2016electric} demonstrates greater effectiveness compared 
to the vanilla model $G_1$ and many other load forecasting models. It mainly selects appropriate lagged hourly temperatures and/or moving average temperatures 
in order to capture the recency effect fully.  The daily moving average temperature of the $b$-th day can be written as:
  \begin{align*} 
  \widetilde{T}_{t,b} = \frac{1}{24} \sum^{24b}_{h=24b-23} T_{t-h}, d = 1,2,3 \cdots,
  \end{align*}
and the Recency model can be expressed as:
  \begin{align*} 
  y_t= & r_0 + r_1 Trend_t + 
  r_2 H_t + r_3 W_t+r_4 M_t \\ & + r_5 H_t W_t + \sum_{h} f(T_{t-h}) + \sum_{d} f(\widetilde{T}_{t,b}),
  \end{align*}

where $h$ and $b$ can be selected by the validation set. Similar to the Recency model, we also incorporate the recency effect variables 
$T_{t-i}, T_{t-i}^2, T_{t-i}^3 , ( 0 \leq i \leq h)$ and 
$\widetilde{T}_{t,j}, \widetilde{T}_{t,j}^2, \widetilde{T}_{t,j}^3, (1 \leq j \leq  d)$ in our HOPS model, 
and we choose the following grid search range:$\{0,1,2,3,\cdots ,24 \}$ 
for $h$ and $\{1, 2, 3,\cdots ,7 \}$ for $b$.
The difference is that we do not include interaction terms as variables so that we take use less and simpler variables.
We denote it as ReHOPS (the Recency HOPS Model).
For practicality and reducing training time, we fix $k_2 = 60$ and $k_3 = 9$. This may not be the optimal choice for $k_2$ and $k_3$, 
but generally, the forecasting performance is not highly sensitive to the embedding dimension. 

We first train the model on the data of 2012 and 2013. Then, we select the optimal h-b
 pair using the data of 2014. Finally, we retrain the model on the data of 2013 and 2014, then forecast on 2015.
 We use two years of data for retraining instead of, as before, three years, in order to maintain consistency with \cite{wang2016electric}.
Both Recency and ReHOPS use the above process to select variables.
The experimental results are presented in Table \ref{table3}. 
It can be observed that our model demonstrates a significant improvement over the Recency model.

Finally, to demonstrate the practicality of the proposed method, we list the total training and forecasting time of 10 datasets for each model as Table \ref{table4}. 

\begin{table}[htbp]
  \centering
  \caption{Comparisons Between the Recency model and the ReHOPS model. }
    \begin{tabular}{m{0.12\textwidth}<{\raggedright} | m{0.07\textwidth}<{\raggedleft} m{0.07\textwidth}<{\raggedleft} | 
      m{0.07\textwidth}<{\raggedleft} m{0.07\textwidth}<{\raggedleft} | m{0.07\textwidth}<{\raggedleft} m{0.07\textwidth}<{\raggedleft} }
    \toprule
    & \multicolumn{2}{c}{} & \multicolumn{2}{c}{} & \multicolumn{2}{c}{ Daily Peak} \\
    & \multicolumn{2}{c}{MAPE(\%) } & \multicolumn{2}{c}{MSE (×10$^3$) } & \multicolumn{2}{c}{MAPE(\%)} \\
    \cmidrule{2-3} \cmidrule{4-5} \cmidrule{6-7}
    Zone & ReHOPS & Recency & ReHOPS & Recency & ReHOPS & Recency\\
    \midrule
    CT             & \textbf{3.01} & 3.31  & \textbf{20.65} & 25.80 & \textbf{2.75} & 3.33  \\
    MASS           & \textbf{2.84} & 3.39  & \textbf{66.00} & 102.16 & \textbf{2.56} & 3.19  \\
    ME             & \textbf{4.50} & 4.84  & \textbf{5.00}  & 6.15  & 3.81 & \textbf{3.73}  \\
    NEMASS         & \textbf{3.20} & 3.60  & \textbf{17.96} & 24.61 & \textbf{3.06} & 3.70  \\
    NH             & \textbf{2.66} & 2.92  & \textbf{2.46}  & 3.10  & \textbf{2.63} & 2.89  \\
    RI             & \textbf{2.89} & 3.28  & \textbf{1.48} & 2.17   & \textbf{2.72} & 3.17  \\
    SEMASS         & \textbf{3.50} & 3.89  & \textbf{6.39} & 8.96   & \textbf{2.82} & 3.50  \\
    VT             & 4.11 & \textbf{4.06}  &  1.09  & \textbf{1.03} & \textbf{3.13} & 3.29  \\
    WCMASS         & \textbf{3.41} & 3.68  & \textbf{8.34} & 10.36  & \textbf{3.08} & 3.51  \\
    \midrule
    \multicolumn{1}{l|}
        {AVERAGE} & \textbf{3.35} & 3.66  & \textbf{14.37} & 20.48 & \textbf{2.95} & 3.37  \\
    \midrule
    TOTAL & \textbf{2.42} & 2.74  & \textbf{220.55} & 313.32 & \textbf{2.30} & 2.66  \\
    \bottomrule
    \end{tabular}
  \label{table3}
\end{table}

\begin{table}[htbp]
  \centering
  \caption{The total time taken by proposed models. }
    \begin{tabular}{m{0.06\textwidth}<{\raggedright} m{0.09\textwidth}<{\raggedleft} m{0.09\textwidth}<{\raggedleft}
      m{0.09\textwidth}<{\raggedleft}m{0.09\textwidth}<{\raggedleft}m{0.09\textwidth}<{\raggedleft}}
      \toprule
    Model & HOPS289 & HOPS47 & HOPS50 & HOPS59 & ReHOPS \\
    \midrule
    Time & 1h6m9s & 19m24s & 20m53s & 26m26s & 2h28m42s \\
    \bottomrule
    \end{tabular}
  \label{table4}
\end{table}

\section{Conclusion}
In this paper, we propose a high-order polynomial forecasting model with self-supervised dimension reduction (HOPS), which results in stable improvements in load forecasting accuracy. Additionally, we introduce a fast algorithm based on FR-CG, 
significantly enhancing the efficiency of solving the numerical optimization problem. 
The results of our numerical experiments also demonstrate that our method can achieve higher forecast accuracy with fewer parameters and less complex variable combinations than its counterparts.

More importantly, our approach can be integrated with many existing load forecasting methods. While numerous existing schemes rely on the 
selection and inclusion of additional influential variables, our method does not depend on variable selection. If additional variables are identified as driving factors of the load, they can be 
incorporated into our proposed framework to further improve forecast accuracy.

\section{Appendix}

\subsection{Appendix \uppercase\expandafter{\romannumeral1}}

\begin{theorem} \label{thm1}
\textbf{Eckart-Young-Mirsky Theorem}
Supposing that $X\in \mathbb{R}^{m\times n}$ is a matrix with $ \textbf{rank}(X) = r$, 
the singular value decomposition (SVD) of $X$ is given by $X=U\Sigma V^T $                             
, where the singular values
$\sigma_1 \geq \sigma_2 \cdots  \geq \sigma_r$ are ordered from largest to smallest.             
Let $k < r$ be an integer, and define the truncated matrix
$X_k = U\Sigma_k V^T$, 
where $\mathbf u_i $
and $\mathbf v_i$ are the $i$-th row vectors of $U$ and $V$, respectively.
$\Sigma_k$ retains the first $k$ singular values of $\Sigma$.
Then for any matrix $B$ with $\textbf{rank}(B) = r$, 
the minimum error in the 2-norm sense is achieved by $X_k$, that is:
\begin{align}
\min_{\textbf{rank} (B) = k} || X-B ||_2 = || X - X_k ||_2 = \sigma_{k+1},
\end{align} 

The same conclusion holds for the Frobenius norm:
\begin{align}
  \min_{\textbf{rank} (B) = k} || X-B ||_F = || X - X_k ||_F 
= \sqrt{\sigma_{k+1}^2 + \cdots + \sigma_r^2},
\end{align} 
\end{theorem}

It is worth noting that if the condition $\textbf{rank} (B) = k$ is replaced by $\textbf{rank} (B) \leq k$, 
we can prove that the above conclusion still holds through Eckart-Young-Mirsky Theorem itself.

\begin{theorem}\label{thm2}
Supposing $\textbf{rank} (X_{m \times n}) = r > k $ and $X_{m \times n} \in \mathbb{R}^{m \times n}$, 
there exists an analytical solution to 
the optimization problem (\ref{opt0}), and the analytical solution is
$ (V^T)^{-1} E_k V^T = V E_k V^T$, where
$V$ is the right singular matrix of $X_{m \times n}$,
$E_k = 
\left[
\begin{array}{ccccccc}
    I_k          & \mathbf{0}\\
    \mathbf{0}   & \mathbf{0}\\
\end{array}
\right]_{n \times n}
$, $I_k$ is an $k$-dimensional identity matrix.

Furthermore, we can obtain $L_{n \times k}$ as follows:
\begin{align} 
L_{n \times k} = V_{n \times n} \left[
    \begin{array}{ccccccc}
        I_k          \\
        \mathbf{0}   \\
    \end{array}
    \right]_{n \times k}.
\end{align} 
\end{theorem}
  
\begin{proof}
First we define the set $\varOmega_1$ and the set $\varOmega_2$:
\begin{align*}
&\varOmega_1 = \{X_{m \times n} D_{n \times n} \in \mathbb{R}^{m \times n} | 
D_{n \times n}=\mathop{\arg\min}_{A_{n \times n} \in \mathbb{R}^{n \times n} , \textbf{rank}(A_{n \times n}) \leq k}
\left\lVert X_{m \times n} A_{n \times n} - X_{m \times n} \right\rVert_F^2 \}, \\
& \varOmega_2 = \{X_k \in \mathbb{R}^{m \times n} | X_k =
\mathop{\arg\min}_{A_{m \times n} \in \mathbb{R}^{m \times n} , \textbf{rank}(A_{m \times n}) \leq k} 
\left\lVert A_{m \times n} - X_{m \times n} \right\rVert_F^2 \}.
\end{align*} 
$\textbf{rank}(X_{m \times n} D_{n \times n}) \leq k$ holds for any
$ D_{n \times n} \in \mathbb{R}^{n \times n} $ with $ \textbf{rank}(D_{n \times n}) \leq k$. 
Therefore, for any $D_{n \times n} $ and $ X_k $ that satisfies 
$ X_{m \times n} D_{n \times n}  \in \varOmega_1$ and $ X_k \in \varOmega_2$
, the following holds:
$$
\left\lVert X_k - X_{m \times n} \right\rVert_F^2 \leq 
\left\lVert X_{m \times n} D_{n \times n} - X_{m \times n} \right\rVert_F^2,
$$
According to Eckart-Young-Mirsky Theorem, there exists 
$U\Sigma_k V^T \in \varOmega_2$. Additionally, let
$$
A_{n \times n} = (V^T)^{-1} E_k V^T,
$$
where $ \textbf{rank}((V^T)^{-1} E_k V^T) = k $.
Thus,
$$
X_{m \times n} A_{n \times n} = (U\Sigma V^T)
((V^T)^{-1} E_k V^T) = (U\Sigma_k V^T),
$$
with $\textbf{rank}(A_{n \times n})=k$, therefore
\begin{align*}
  \left\lVert X_k - X_{m \times n} \right\rVert_F^2 =
\left\lVert X_{m \times n}A_{n \times n} - X_{m \times n} \right\rVert_F^2 \\
\geq \left\lVert X_{m \times n}D_{n \times n} - X_{m \times n} \right\rVert_F^2.
\end{align*}
This implies that for any 
$ X_{m \times n} D_{n \times n} \in \varOmega_1$ and $ X_k \in \varOmega_2$, 
the following equality holds:
$$
\left\lVert X_k - X_{m \times n} \right\rVert_F^2 =
\left\lVert X_{m \times n} D_{n \times n} - X_{m \times n} \right\rVert_F^2.
$$
Since $V$ is an orthogonal matrix and $(V^T)^{-1}=V$ ,$D_{n \times n} $ $ = V E_k V^T = (V^T)^{-1} E_k V^T$
is an analytical solution to the optimization problem (\ref{opt0}).
To obtain $L_{n \times k}$, we take the left half of $V E_k V^T$,
which is:
\begin{align*}
L_{n \times k} = V_{n \times n} \left[
    \begin{array}{ccccccc}
        I_k          \\
        \mathbf{0}   \\
    \end{array}
    \right]_{n \times k}.
\end{align*} 
In other words, we can obtain the linear transformation matrix $L_{n \times k}$
by selecting the first $k$ columns of $V$.
It is not hard to extend the above conclusion to the condition $\textbf{rank} (A_{m \times n}) = k$ 
by Eckart-Young-Mirsky Theorem because 
$\min_{\textbf{rank} (B) = k} || X-B ||_F = \min_{\textbf{rank} (B) \leq k} || X-B ||_F $ .
\end{proof}

\subsection{Appendix \uppercase\expandafter{\romannumeral2}}
\begin{theorem}\label{thm3}
The optimal solution to optimization problem (\ref{lq3}) exists and this solution 
is either unique or an affine space of $\mathbb{R}^{\sum_{i = 0}^d (k_i)^i}$ with infinitely many elements.
\end{theorem}

\begin{proof}

  First, we summarize the core idea of the proof. From the expression in 
  (\ref{multipolynomial2}) for an embedded $n$-dimensional $d$-degree polynomial model, 
  it is evident that the predicted value
  $\widetilde{f} \left(x_{1\times n},\boldsymbol{W_0},(\boldsymbol{W_1})_{k_1}, \cdots, (\boldsymbol{W_d})_{k_d}\right)$
  is linear with respect to any of the polynomial parameters.
  In other words, If we expand all the polynomial parameters 
  from tensor form $(\boldsymbol{W}_i)_{k_i}$ $(i = 1,2,\cdots d)$ into a vector form 
  $\boldsymbol{\widehat{w}} $, we obtain a very large-scale linear regression problem.

  Thus, compared with MLR, the embedded $d$-degree polynomial model can be interpreted as a regression model in 
  which the number of parameters increases from $n+1$ (where $n$ is the input data 
  dimension) to $\sum^{d}_{i = 0} (k_i)^i$, given the same input feature matrix $X_{m \times n}$,
  where $X_{m \times n}$ is the matrix obtained by concatenating the sample vectors
   $x_{1\times n}$ in rows. 
   Essentially, we treat the interactions as new input variables.

  Specifically, we can flatten the $i$-degree monomial parameter tensor 
  $(\boldsymbol{W_i})_{k_i}$ into a row vector denoted as
  $\widehat{\boldsymbol{w}}_i$, which becomes a $(k_i)^i$-dimensional vector.
  We concatenate all the $i$-degree monomial parameters together to obtain
  $$
  \widehat{\boldsymbol{w}} =
  (\widehat{\boldsymbol{w}}_1, \widehat{\boldsymbol{w}}_2, \cdots, \widehat{\boldsymbol{w}}_d) ,
  $$
  Similarly, we flatten
  $\underbrace{(\boldsymbol{x}\cdot L_{n \times k} )\otimes (\boldsymbol{x}\cdot L_{n \times k} ) \otimes \cdots (\boldsymbol{x}\cdot L_{n \times k} )}_i $
  into a row vector $\widehat{\boldsymbol{x}}_i (i=1,2,\cdots d)$ 
  and concatenate them as follows:
  $$
  \widehat{\boldsymbol{x} } =
  (\widehat{\boldsymbol{x}}_0, \widehat{\boldsymbol{x}}_1, \cdots, \widehat{\boldsymbol{x}}_d).
  $$
  Now, the model can be formulated as a linear regression fitting problem:
  \begin{align*}
    \widetilde{f}( x_{1\times n} )= \widehat{\boldsymbol{w}} \cdot \widehat{\boldsymbol{x} }^T,
  \end{align*}
  This demonstrates that solving the polynomial model is equivalent to solving the 
  following optimization problem:
  \begin{align*}
       \min_{\widehat{\boldsymbol{w}} \in \mathbb{R}^{\sum_{i = 0}^d (k_i)^i}} 
       || \widehat{\boldsymbol{X} } \cdot\widehat{\boldsymbol{w}}^T - Y ||_2^2,
  \end{align*}
  where $\widehat{\boldsymbol{X}}$ represents a matrix concatenated row-wise by sample 
  row vectors $\widehat{\boldsymbol{x}}$ and 
  $Y$ denotes all the labels (a column vector). 
  The solution to this problem is either unique or an affine space of $\mathbb{R}^{\sum_{i = 0}^d (k_i)^i}$ with infinitely many elements, 
  which depends on whether $ \widehat{\boldsymbol{X} } \cdot\widehat{\boldsymbol{w}}^T = Y $ is well-determined, 
  under-determined or over-determined.
\end{proof}

\subsection{Appendix \uppercase\expandafter{\romannumeral3}}
\begin{theorem} \label{thm4}
  Supposing that $\textbf{rank}(X_{m\times n}) > k_i,i=1,2,\cdots,d$ and for all $k_i$ (the embedding dimension of $i$-order terms), the solution to 
  optimization problem (\ref{opt0}) is $D_{n \times n}^i$ 
  $(\textbf{rank}(D_{n \times n}^i) = k_i, i = 1,2,\cdots d)$. For any 
  $$
  L_{n \times k_i} \in \mathbb{R}^{n \times k_i}, R_{k_i \times n} \in \mathbb{R}^{k_i \times n} , i = 1,2,\cdots d,
  $$ and 
  $$
  L_{n \times k_i}' \in \mathbb{R}^{n \times k_i}, R_{k_i \times n}' \in \mathbb{R}^{k_i \times n} , i = 1,2,\cdots d,
  $$
  which satisfy
  $$
  D_{n \times n}^i = L_{n \times k_i} R_{k_i \times n} = L_{n \times k_i}' R_{k_i \times n}'  , i = 1,2,\cdots d,
  $$
  and there exists $i$ such that the following inequality holds
  $$
  L_{n \times k_i} \neq L_{n \times k_i}', R_{k_i \times n} \neq R_{k_i \times n}',
  $$
  the following conclusion can be drawn:
  $L_{n \times k_i} $ and $ L_{n \times k_i}'$ ,$ i = 1,2,\cdots d $,
  are equivalent in our model (\ref{multipolynomial2}). In other words, for any
  $ \boldsymbol{W_0},(\boldsymbol{W_1})_{k_1},\cdots (\boldsymbol{W_d})_{k_d}$ ,
  there exist 
  $\boldsymbol{W_0'},(\boldsymbol{W_1'})_{k_1},\cdots (\boldsymbol{W_d'})_{k_d}$,
  such that the following equation holds for any $\boldsymbol{x} \in \mathbb{R}^n$:
  \begin{align*} 
  & \widetilde{f}' \left(\boldsymbol{x}, \boldsymbol{W_0'},(\boldsymbol{W_1'})_{k_1},\cdots (\boldsymbol{W_d'})_{k_d} \right)
  \\ & = \textbf{sum}(\boldsymbol{W_0'}*X_0') + \textbf{sum}((\boldsymbol{W_1'})_{k_1}*X_1') \cdots +
  \textbf{sum} ((\boldsymbol{W_d'})_{k_d}*X_d')
  \\ & = \textbf{sum}(\boldsymbol{W_0}*X_0) + \textbf{sum}((\boldsymbol{W_1})_{k_1}*X_1) \cdots +
  \textbf{sum} ((\boldsymbol{W_d})_{k_d}*X_d),
  \\ & = \widetilde{f} \left(\boldsymbol{x}, \boldsymbol{W_0},(\boldsymbol{W_1})_{k_1},\cdots (\boldsymbol{W_d})_{k_d} \right) ,
  \end{align*}
  where $X_i = \underbrace{\boldsymbol{x}L_{n \times k_i} \otimes \boldsymbol{x}L_{n \times k_i} \cdots \otimes \boldsymbol{x}L_{n \times k_i}}_i$
  ,$X_i' = \underbrace{\boldsymbol{x}L_{n \times k_i}' \otimes \boldsymbol{x}L_{n \times k_i}' \cdots \otimes \boldsymbol{x}L_{n \times k_i}' }_i, i = 1,2,\cdots d $
  and $\boldsymbol{x} = x_{1\times n}$.
  \end{theorem}

\begin{proof}
\par Supposing that there is a $L_{n \times k}$ and a $L_{n \times k}^{\prime}$ satisfy the hypothesis and 
$L_{n \times k} \neq  L_{n \times k}^{\prime}$. 
We first consider the case of the $n$-dimensional $2$-degree polynomial:
\begin{align*}
\widetilde{f} \left(\boldsymbol{x}, \boldsymbol{W_2}, \boldsymbol{W_1}, \boldsymbol{W_0} \right) = \boldsymbol{x}
\boldsymbol{W_2}\boldsymbol{x}^T + \boldsymbol{W_1}\boldsymbol{x}^T + \boldsymbol{W_0},
\\ \quad \boldsymbol{W_2} \in 
\mathbb{R}^{n \times n},\boldsymbol{W_1} \in \mathbb{R}^{n} ,\boldsymbol{W_0} \in \mathbb{R}.
\end{align*}
When the input $\boldsymbol{x}$ is compressed by $L_{n \times k}$, then:
\begin{align*}
& \widetilde{f} \left(\boldsymbol{x}, (\boldsymbol{W_2})_{k_2}, (\boldsymbol{W_1})_{k_1}, \boldsymbol{W_0} \right) 
\\  = & \left(\boldsymbol{x}L_{n \times k_2}\right)
(\boldsymbol{W_2})_{k_2}\left(L_{n \times k_2}^T \boldsymbol{x}^T\right) 
+ (\boldsymbol{W_1})_{k_1}\left(L_{n \times k_1}^T \boldsymbol{x}^T\right) + \boldsymbol{W_0} \\
 = & \boldsymbol{x} \left(L_{n \times k_2}
(\boldsymbol{W_2})_{k_2} L_{n \times k_2}^T\right)_{n \times n }  \boldsymbol{x}^T
+ \left((\boldsymbol{W_1})_{k_1} L_{n \times k_1}^T\right)_{n}  \boldsymbol{x}^T + \boldsymbol{W_0} ,\\
& (\boldsymbol{W_2})_{k_2} \in \mathbb{R}^{k_2 \times k_2}, (\boldsymbol{W_1})_{k_1} \in \mathbb{R}^{k_1}, \boldsymbol{W_0} \in \mathbb{R}.
\end{align*} 
In the same manner:
\begin{align*}  
& \widetilde{f} \left(\boldsymbol{x}, (\boldsymbol{W_2'})_{k_2}, 
(\boldsymbol{W_1'})_{k_1}, \boldsymbol{W_0'} \right) 
\\ = & \boldsymbol{x} \left(L_{n \times k_2}^{\prime}
(\boldsymbol{W_2'})_{k_2} (L_{n \times k_2}^{\prime})^T\right)_{n \times n }  \boldsymbol{x}^T
+ \left((\boldsymbol{W_1'})_{k_1} (L_{n \times k_1}^{\prime})^T\right)_{n}  \boldsymbol{x}^T
\\ & + \boldsymbol{W_0'}, \\
& (\boldsymbol{W_2'})_{k_2} \in \mathbb{R}^{k_2 \times k_2}, (\boldsymbol{W_1'})_{k_1} \in \mathbb{R}^{k_1}, \boldsymbol{W_0'} \in \mathbb{R}.
\end{align*} 

Actually, we just need to show that, for any $(\boldsymbol{W_2})_{k_2} $ and 
$(\boldsymbol{W_1})_{k_1}$, there exists $(\boldsymbol{W_2'})_{k_2}$ and
$(\boldsymbol{W_1'})_{k_1}$ such that:
\begin{align*}
\left(L_{n \times k_2}^{\prime}(\boldsymbol{W_2'})_{k_2} (L_{n \times k_2}^{\prime})^T\right)_{n \times n } = 
\left(L_{n \times k_2}(\boldsymbol{W_2})_{k_2} L_{n \times k_2}^T\right)_{n \times n },
\end{align*} 
and
\begin{align*}
    \left((\boldsymbol{W_1'})_{k_1} (L_{n \times k_1}^{\prime})^T\right)_{n} = 
    \left((\boldsymbol{W_1})_{k_1} L_{n \times k_1}^T\right)_{n},
\end{align*} 
and then, through the arbitrariness of $L_{n \times k}$ and $L_{n \times k}^{\prime}$ we can prove 
that they are equivalent under different embedding matrices.

We now prove this claim.

First, without loss of generality, denote $k_i$ as $k$.
Since $L_{n \times k}R_{k \times n}=  D_{n \times n}$ and $\textbf{rank}(D_{n \times n}) = k$,
, it follows that $\textbf{rank}(R_{k \times n}) = k$.
According to the basis expansion theorem, we can always find another $(n-k)$ basis in $\mathbb{R}^n$
independent of the $k$ row vectors of $R_{k \times n}$. We arrange them in turn and
obtain a matrix $C_{(n-k) \times n}$ with full row rank. Merging $R_{k \times n}$ and 
$C_{(n-k) \times n}$ by row, we obtain the following:
\begin{align*}
H = \left[
\begin{array}{cc}
R_{k \times n}   \\
C_{(n-k) \times n} \\
\end{array}
\right]_{n \times n},
\end{align*} 
where $H$ is invertible. Let the first $k$ columns of $H^{-1}$ be 
$\widetilde{R}_{n \times k}$,  and the last $(n-k)$ columns be 
$\widetilde{C}_{n \times (n-k)}$. Hence 
\begin{align*}
& \left[
\begin{array}{cc}
R_{k \times n}   \\
C_{(n-k) \times n} \\
\end{array}
\right]_{n \times n}
\left[
\begin{array}{cc}
    \widetilde{R}_{n \times k} & \widetilde{C}_{n \times (n-k)}  \\
\end{array}
\right]_{n \times n} \\ & = \left[
    \begin{array}{cc}
    I_k  & 0 \\
    0    & I_{(n-k)} \\
    \end{array}
    \right]_{n \times n} = I_n,
\end{align*} 
Where $I_n, I_k, I_{(n-k)}$ are $n$-dimensional, $k$-dimensional and $(n-k)$-dimensional unit matrices respectively. 
Thus,
\begin{align*}
    R_{k \times n} \widetilde{R}_{n \times k} = I_k.
\end{align*} 
Since 
\begin{align*}
    L_{n \times k}R_{k \times n} =D_{n \times n} 
    =  L_{n \times k}^{\prime} R_{k \times n}^{\prime},
\end{align*} 
and 
\begin{align*}
    &L_{n \times k}R_{k \times n} \widetilde{R}_{n \times k}
    = D_{n \times n}\widetilde{R}_{n \times k} 
    = L_{n \times k}^{\prime} R_{k \times n}^{\prime}\widetilde{R}_{n \times k}, 
\end{align*} 
it follows that:
\begin{align*}
    & L_{n \times k}
    = L_{n \times k}^{\prime} R_{k \times n}^{\prime}\widetilde{R}_{n \times k}.
\end{align*} 
Thus, setting $k = k_2,$ we obtain:
\begin{align*}
    & L_{n \times k_2} (\boldsymbol{W_2})_{k_2} L_{k_2 \times n} ^T \\
    & = L_{n \times k_2}^{\prime} R_{k_2 \times n}^{\prime}\widetilde{R}_{n \times k_2}
    (\boldsymbol{W_2})_{k_2}
    (L_{n \times k_2}^{\prime} R_{k_2 \times n}^{\prime}\widetilde{R}_{n \times k_2})^T \\
    & = L_{n \times k_2}^{\prime} \underbrace{R_{k_2 \times n}^{\prime}\widetilde{R}_{n \times k_2}
    (\boldsymbol{W_2})_{k_2}
    \widetilde{R}_{k_2 \times n}^T (R^{\prime})^T_{n \times k_2}}_{(\boldsymbol{W_2'})_{k_2}}
    (L^{\prime})^T_{k_2 \times n}\\
    & = L_{n \times k_2}^{\prime} (\boldsymbol{W_2'})_{k_2} (L^{\prime})^T_{k_2 \times n}.
\end{align*} 
In a similar manner, we can show:
\begin{align*}
    (\boldsymbol{W_1})_{k_1} L_{n \times k_1}^T
    & = (\boldsymbol{W_1})_{k_1}
    (L_{n \times k_1}^{\prime} R_{k_1 \times n}^{\prime}\widetilde{R}_{n \times k_1})^T \\
    & = \underbrace{(\boldsymbol{W_1})_{k_1} 
    \widetilde{R}_{k_1 \times n}^T (R^{\prime})^T_{n \times k_1}}_{(\boldsymbol{W_1'})_{k_1}}
    (L^{\prime})^T_{k_1 \times n}\\
    & = (\boldsymbol{W_1'})_{k_1} (L^{\prime})^T_{k_1 \times n}.
\end{align*} 
So far, we have proved the result for the case of $n$-dimensional $2$-order polynomial, 
and in the following we extend this conclusion to higher degrees.

To illustrate, take the  $n$-dimensional $3$-order polynomial as an example.
The case of higher degrees can be proved similarly.

We first consider the form in expression (\ref{multipolynomial2}):
\begin{align*} 
& \widetilde{f} \left(\boldsymbol{x}, \boldsymbol{W_0},(\boldsymbol{W_1})_{k_1},(\boldsymbol{W_2})_{k_2}, (\boldsymbol{W_3})_{k_3} \right) =
\textbf{sum}(\boldsymbol{W_0}*X_0) + 
\\ &  \textbf{sum}((\boldsymbol{W_1})_{k_1}*\widetilde{X} _1) + \textbf{sum}((\boldsymbol{W_2})_{k_2}*\widetilde{X} _2)
+ \textbf{sum}((\boldsymbol{W_3})_{k_3}*\widetilde{X} _3),
\end{align*}
To make the proof clearer, we may as well denote $(\boldsymbol{W}_3)_{k_3}$ and
 $(\boldsymbol{W}_3')_{k_3}$ as $(\boldsymbol{W}_3)_{k_3 \times k_3 \times k_3}$ and
 $(\boldsymbol{W}_3')_{k_3 \times k_3 \times k_3}$.
Since the conclusion for the primary term and quadratic term has already been proved, we only need to 
consider the cubic term and just prove for any 
$(\boldsymbol{W}_3)_{k_3 \times k_3 \times k_3}$, there is
$(\boldsymbol{W}_3^{\prime})_{k_3 \times k_3 \times k_3}$, such that
\begin{align} \label{equal}
& \textbf{sum}((\boldsymbol{W}_3)_{k_3 \times k_3 \times k_3}*(\boldsymbol{x}L_{n \times k_3}\otimes \boldsymbol{x}L_{n \times k_3}
\otimes \boldsymbol{x}L_{n \times k_3}) )
\notag \\ = & 
\textbf{sum}( (\boldsymbol{W}_3^{\prime})_{k_3 \times k_3 \times k_3}*(\boldsymbol{x}L_{n \times k_3}^{\prime}\otimes \boldsymbol{x}L_{n \times k_3}^{\prime}
\otimes \boldsymbol{x}L_{n \times k_3}^{\prime}) ),
\end{align}
holds for any $\boldsymbol{x} \in \mathbb{R}^n$.

For more convenient representation, we utilize Einstein notation to represent operations, and then we can obtain:
\begin{align*} 
&\textbf{sum}( (\boldsymbol{W}_3)_{k_3 \times k_3 \times k_3}*(\boldsymbol{x}L_{n \times k_3}\otimes \boldsymbol{x}L_{n \times k_3}
\otimes \boldsymbol{x}L_{n \times k_3}) ) \\
=& \textbf{sum}( (L_{n \times k_3})_{ai}(L_{n \times k_3})_{bj}(L_{n \times k_3})_{ch}
((\boldsymbol{W}_3)_{k\times k_3\times k_3})^{ijh}
*(\boldsymbol{x}
\otimes \boldsymbol{x}\otimes \boldsymbol{x}) ) \\
=& \textbf{sum}( (L_{n \times k_3}^{\prime} R_{k \times n}^{\prime}\widetilde{R}_{n \times k_3})_{ai}
(L_{n \times k_3}^{\prime} R_{k_3 \times n}^{\prime}\widetilde{R}_{n \times k_3})_{bj}
(L_{n \times k_3}^{\prime} R_{k_3 \times n}^{\prime}\widetilde{R}_{n \times k_3})_{ch}
((\boldsymbol{W}_3)_{k_3 \times k_3 \times k_3})^{ijh}*(\boldsymbol{x}
\otimes \boldsymbol{x}\otimes \boldsymbol{x}) ) \\
=& \textbf{sum}( (L_{n \times k_3}^{\prime})_{ad}
(L_{n \times k_3}^{\prime})_{be}
(L_{n \times k_3}^{\prime})_{cf} 
(
(R_{k_3 \times n}^{\prime}\widetilde{R}_{n \times k_3})_{di}
(R_{k_3 \times n}^{\prime}\widetilde{R}_{n \times k_3})_{ej}
(R_{k_3 \times n}^{\prime}\widetilde{R}_{n \times k_3})_{fh}
((\boldsymbol{W}_3)_{k_3 \times k_3 \times k_3})^{ijh} )^{def}
* (\boldsymbol{x}
\otimes \boldsymbol{x}\otimes \boldsymbol{x}))
\\ = & \textbf{sum}( (
(R_{k_3 \times n}^{\prime}\widetilde{R}_{n \times k_3})_{di}
(R_{k_3 \times n}^{\prime}\widetilde{R}_{n \times k_3})_{ej}
(R_{k_3 \times n}^{\prime}\widetilde{R}_{n \times k_3})_{fh}
((\boldsymbol{W}_3)_{k_3 \times k_3 \times k_3})^{ijh} )
* (\boldsymbol{x}L_{n \times k_3}^{\prime}
\otimes \boldsymbol{x}L_{n \times k_3}^{\prime} 
\otimes \boldsymbol{x}L_{n \times k_3}^{\prime})).
\end{align*}
We only need to let 
\begin{align*} 
& (\boldsymbol{W}_3^{\prime})_{k_3 \times k_3 \times k_3} 
\\ = &
(
(R_{k_3 \times n}^{\prime}\widetilde{R}_{n \times k_3})_{di}
(R_{k_3 \times n}^{\prime}\widetilde{R}_{n \times k_3})_{ej}
(R_{k_3 \times n}^{\prime}\widetilde{R}_{n \times k_3})_{fh}
((\boldsymbol{W}_3)_{k_3 \times k_3 \times k_3})^{ijh} ),
\end{align*}
and then Equation (\ref{equal}) holds for any $\boldsymbol{x} \in \mathbb{R}^n$. The generalization 
to higher degrees can be deduced in a similar manner.
\end{proof}
  
To make the proof easier to understand, we briefly explain the practical 
significance of Einstein notation.

Einstein notation is a compact way to express tensor operations, 
generalizing matrix products. Supposing the tensors
$A_{l_1 \times l_2 \times n} \in \mathbb{R}^{l_1 \times l_2 \times n}$ and 
$B_{m_1 \times m_2 \times n} \in \mathbb{R}^{m_1 \times m_2 \times n}$, we can obtain that 
\begin{align*}
& \underbrace{\left( (A_{l_1 \times l_2 \times n})_{aci}(B_{m_1 \times m_2 \times n})^{bdi}\right)
}_{l_1 \times l_2 \times m_1 \times m_2} \left[p,q,r,s \right]
\\ & = \sum_{i=1}^{k}
(A_{l_1 \times l_2 \times n})\left[p,q,i\right](B_{m_1 \times m_2 \times n})\left[r,s,i \right],
\end{align*}   

where the indices $\left[p,q,r,s \right]$ denotes the (p,q,r,s)-th element of the 
tensor, and the brackets follow the same meaning throughout.

To illustrate this with a simpler example, we consider a $2$-order tensor (a matrix). 
Supposing there are 3 matrices: $A \in \mathbb{R}^{l \times n}$ , $B \in \mathbb{R}^{m \times n}$ 
and $H \in \mathbb{R}^{n \times n}$, it follows that
\begin{align*}
\underbrace{(A_{l \times n})_{ai}(B_{m \times n})_{bj}(H_{n \times n})^{ij}}_{l \times m} 
= \underbrace{(A_{l \times n}H_{n \times n}(B^T)_{n \times m})}_{l \times m}.
\end{align*}   
In this case, matrix $B_{m \times n}$ is multiplied and then summed according to the second 
axis of $H_{n \times n}$ to obtain a new matrix. Similarly, matrix $A_{l \times n}$ is 
multiplied and then summed according to the first aixs of $H_{n \times n}$. Therefore, 
this expression in Einstein notation is equivalent to matrix multiplication.

In this context, using Einstein notation, we can express:
\begin{align*}
  &\textbf{sum} \left( (\boldsymbol{W}_2)_{k \times k}*(\boldsymbol{x}L_{n \times k}
\otimes \boldsymbol{x}L_{n \times k}) \right)\\
= & ( \boldsymbol{x}L_{n \times k}
    (\boldsymbol{W}_2)_{k \times k}
    (L^T)_{k \times n} \boldsymbol{x}^T ) \\
= & \textbf{sum} ( L_{n \times k}(\boldsymbol{W}_2)_{k \times k}
    (L^T)_{k \times n} *(\boldsymbol{x}\otimes \boldsymbol{x}) ) \\
= & \textbf{sum} ( L_{n \times k}^{\prime} R_{k \times n}^{\prime}\widetilde{R}_{n \times k}
    (\boldsymbol{W}_2)_{k \times k}
    \widetilde{R}_{k \times n}^T (R^{\prime})^T_{n \times k}(L^{\prime})^T_{k \times n}
    *(\boldsymbol{x}\otimes \boldsymbol{x}) ) \\
= & \textbf{sum} ( (L_{n \times k}^{\prime})_{ac}(L_{n \times k}^{\prime})_{bd}
((R_{k \times n}^{\prime}\widetilde{R}_{n \times k})_{ci}
(R_{k \times n}^{\prime}\widetilde{R}_{n \times k})_{dj}
(\boldsymbol{W}_2)_{k \times k}^{ij})^{cd} *(\boldsymbol{x}
\otimes \boldsymbol{x}). )
\end{align*}   

Thus, it's evident that the matrix form of the $n$-dimensional $2$-order polynomial fits into this unified form.
If readers wish to verify this process, they can use the $torch.einsum()$
function from the PyTorch framework. PyTorch not only serves as an excellent 
deep learning framework but also offers significant advantages in tensor processing.

\bibliographystyle{unsrtnat}
\bibliography{refer.bib}

\end{document}